\documentclass[sn-basic]{sn-jnl}

\usepackage{graphicx}%
\usepackage{multirow}%
\usepackage{amsmath,amssymb,amsfonts}%
\usepackage{amsthm}%
\usepackage{mathrsfs}%
\usepackage[title]{appendix}%
\usepackage{xcolor}%
\usepackage{textcomp}%
\usepackage{manyfoot}%
\usepackage{booktabs}%
\usepackage{algorithm}%
\usepackage{algorithmicx}%
\usepackage{algpseudocode}%
\usepackage{listings}%

\newcommand{\N}{$\mathrm{N}$}
\newcommand{\Ca}{$\mathrm{C\alpha}$}
\newcommand{\C}{$\mathrm{C}$}
\newcommand{\SO}[1]{$\mathrm{SO}(#1)$}
\newcommand{\so}[1]{$\mathfrak{so}(#1)$}

\theoremstyle{thmstyleone}%
\theoremstyle{thmstyletwo}%

\theoremstyle{thmstylethree}%

\begin{document}

\title[Representation choice shapes the interpretation of protein conformational dynamics]{Representation choice shapes the interpretation of protein conformational dynamics}

\author*[1,2]
    {\fnm{Giottonini} \sur{Axel}}
    \email{axel.giottonini@unibe.ch}

\author*[1,3]
    {\fnm{Lemmin} \sur{Thomas}}
    \email{thomas.lemmin@unibe.ch}

\affil*[1]{
    \orgdiv{Institute of Biochemistry and Molecular Medicine},
    \orgname{University of Bern},
    \orgaddress{
        \street{Bühlstrasse 28},
        \city{Bern},
        \postcode{3012},
        \state{Bern},
        \country{Switzerland}
    }
}

\affil[2]{
    \orgdiv{Graduate School for Cellular and Biomedical Sciences (GCB)},
    \orgname{University of Bern},
    \orgaddress{
        \street{Mittelstrasse 43},
        \city{Bern},
        \postcode{3012},
        \state{Bern},
        \country{Switzerland}
    }
}

\affil[3]{
    \orgdiv{Department of Digital Medicine},
    \orgname{University of Bern},
    \orgaddress{
        \street{Murtenstrasse 31},
        \city{Bern},
        \postcode{3008},
        \state{Bern},
        \country{Switzerland}
    }
}

\abstract{
Molecular dynamics simulations provide detailed trajectories at the atomic level, but extracting interpretable and robust insights from these high-dimensional data remains challenging. In practice, analyses typically rely on a single representation. Here, we show that representation choice is not neutral: it fundamentally shapes the conformational organization, similarity relationships, and apparent transitions inferred from identical simulation data.

To complement existing representations, we introduce Orientation features, a geometrically grounded, rotation-aware encoding of protein backbone. We compare it against common descriptions across three dynamical regimes: fast-folding proteins, large-scale domain motions, and protein-protein association. Across these systems, we find that different representations emphasize complementary aspects of conformational space, and that no single representation provides a complete picture of the underlying dynamics.

To facilitate systematic comparison, we developed ManiProt, a library for efficient computation and analysis of multiple protein representations. Our results motivate a comparative, representation-aware framework for the interpretation of molecular dynamics simulations.
}


\maketitle

\section{Introduction}\label{sec1}

Molecular Dynamics (MD) simulations provide a computational framework for resolving the structural and dynamical behavior of biomolecules at atomic resolution. By integrating classical equations of motion, MD enables the study of protein thermodynamics and kinetics \cite{LindorffLarsen2011}, the exploration of conformational landscapes \cite{Zimmerman2021,Chen2022}, and the characterization of protein--ligand binding mechanisms \cite{Morris2021}. Advances in hardware and algorithms have expanded accessible simulation timescales from nanoseconds to hundreds of microseconds and beyond. In particular, GPU acceleration \cite{Pande2010,Voelz2023} and specialized architectures such as Anton \cite{Shaw2014,Shaw2021} now permit routine generation of trajectories that capture folding, binding, and allosteric transitions. As a result, MD simulations have become a central source of mechanistic insight across structural biology, biophysics, and drug discovery.

Despite these advances, extracting biologically meaningful information from raw atomic trajectories remains a fundamental challenge. MD simulations generate high-dimensional, strongly correlated time series, in which functionally relevant motions are often subtle, collective, and distributed across many degrees of freedom. Processes such as allosteric communication, enzymatic rearrangements, and domain motions in molecular machines cannot be directly inferred from Cartesian coordinates alone, and require appropriate representations to reveal their underlying structure and timescales.

A key but often implicit choice in MD analysis is the representation of molecular motion. Dimensionality-reduction, clustering, and kinetic modeling methods typically operate in Euclidean vector spaces, using atomic coordinates, pairwise distances, or torsion angles as input features \cite{Glielmo2021}. Although widely used, these representations impose simplifying assumptions that obscure the nonlinear geometry and rotational constraints inherent to protein conformational space. Recent work has sought to address this limitation by formulating molecular motion on Riemannian manifolds \cite{Diepeveen2023}. In parallel, developments in geometric deep learning have highlighted the importance of \SO{3}-equivariance, ensuring that molecular descriptions are invariant to global rotation, often through the use of residue-level local coordinate systems (LCSs) \cite{Ingraham2019,Jumper2021,Satorras2021,Klein2023}.

Here, we introduce \emph{Orientation features}, a geometrically grounded representation for analyzing protein backbone dynamics in MD simulations. \emph{Orientation features} describes the motion of the backbone as time-dependent rotations of local reference frames of residues, represented in the Lie group \SO{3} and mapped to its Lie algebra \so{3}. This formulation yields a compact, rotation-aware description of conformational change that is naturally aligned with protein geometry. Using a benchmark set of 23 long MD trajectories from D.\,E.\ Shaw Research \cite{LindorffLarsen2011,Pan2019,Chen2022,deshawcovid}, we compare \emph{Orientation features} to commonly used molecular representations, including Cartesian coordinates, torsion angles, and Pointcloud-based features. We find that dynamical signatures extracted from identical simulations strongly depend on the chosen representation, with different representations emphasizing distinct aspects of motion and timescale separation. No single representation consistently captures all relevant dynamical features across systems. To support systematic exploration of these effects, we developed ManiProt, an open-source library, for computing and comparing protein dynamics across multiple coordinate representations.

\section{Results}\label{sec2}

\begin{figure}
	\centering
    \includegraphics[width=\textwidth]{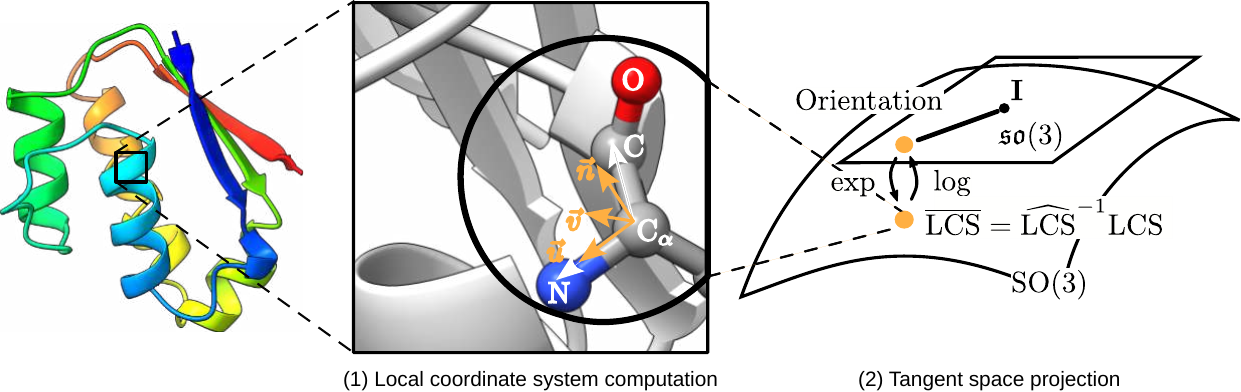}
    \caption{
        \textbf{Orientation Features Workflow.}
        (\textbf{1}) A Local Coordinate System is constructed for each residue from the amide nitrogen (N), alpha carbon (C$\alpha$), and carbonyl carbon (C) positions, forming an orthonormal basis.
        (\textbf{2}) Each LCS is aligned to the per-residue intrinsic mean orientation and projected onto the tangent space at the identity element of SO(3), yielding features in the Lie algebra $\mathfrak{so}(3)$.
    }
    \label{fig:figure_1}
\end{figure}

A central challenge in analyzing MD trajectories is the construction of features that are both compact and geometrically meaningful. To address this challenge, we introduce \emph{Oritentation Features}, a geometry-aware representation of protein backbone dynamics-based on residue-level Local Coordinate Systems (LCSs). Each residue is associated with a local reference frame defined by its backbone atoms (\N, \Ca, \C), allowing backbone motion to be described in terms of local rotations rather than absolute atomic displacements \cite{Ingraham2019,Jumper2021}. After alignment to a reference structure to remove global rigid-body motion, trajectory snapshots are mapped to \emph{Orientation features} through projection onto a tangent space (Figure~\ref{fig:figure_1}). This representation emphasizes rotational degrees of freedom and provides a compact encoding of backbone motion that is sensitive to both local conformational fluctuations and larger-scale collective rearrangements. We evaluate this representation across three complementary dynamical regimes: fast-folding proteins, large-scale domain motions in a multi-domain helicase, and interface rearrangements during protein--protein association.

\subsection{Orientation Features: Mathematical formulation}

We formalize \emph{Orientation features} by modeling protein backbone conformations as elements of a product of rotation groups. Each residue-local coordinate system is represented by an orthonormal matrix in the special orthogonal group \SO{3}. A protein with $n$ residues is therefore described as a point in the product manifold $\mathrm{SO}(3)^n$, where each factor corresponds to the orientation of a single residue-local frame \cite{Berger2003}.

This product structure induces a decoupled geometric representation in which each residue contributes independently to the total configuration. As a result, the associated metric does not explicitly encode correlations between neighboring residues. This simplifying assumption allows for a tractable and interpretable representation of backbone motion, while inter-residue correlations are recovered implicitly through analyses performed on the joint distribution of residue orientations across the trajectory.

Global rotational invariance is handled by working modulo rigid-body rotations. Formally, this corresponds to the quotient space $\mathrm{SO}{3}^n / \mathrm{SO}{3}$, in which conformations differing only by a global rotation are identified. In practice, we realize this quotient by aligning each trajectory frame to a common reference structure, ensuring that \emph{Orientation features} capture only internal conformational changes.

To obtain a linear representation suitable for downstream analysis, each residue-local rotation is mapped from $\mathrm{SO}{3}$ to its Lie algebra $\mathrm{so}{3}$ using the logarithmic map. The Lie algebra $\mathfrak{so}{3}$ can be identified with $\mathbb{R}^3$ via the axis--angle representation, in which the direction of the vector specifies the rotation axis and its norm corresponds to the rotation angle. The logarithmic map is well-defined for rotations with angles less than $\pi$; for rare cases in which rotations approach this limit and numerical instabilities arise, we employ a stabilized approximation described in the Methods.

Following alignment, \emph{Orientation features} are constructed by expressing each residue-local frame relative to a reference orientation $\widehat{\mathrm{LCS}}$ via left multiplication, $\widehat{\mathrm{LCS}}^{-1} \cdot \mathrm{LCS}$. This step centers rotations around the identity, enabling consistent application of the logarithmic map. The logarithm is then applied independently to each corrected frame, yielding a $3n$-dimensional Euclidean feature vector in which each residue contributes three rotational degrees of freedom.

We consider two variants of this representation that differ in the choice of reference frame. In the first, denoted Orientation, a single reference structure is used, typically corresponding to the experimental or initial conformation. This variant emphasizes deviations relative to a specific structural state. In the second variant, denoted Orientation~$\odot$, the reference frame for each residue is chosen as the intrinsic mean of its local coordinate systems over the trajectory \cite{Fletcher2004}. This residue-specific reference yields a representation that captures rotational variability across the full conformational ensemble. The symbol $\odot$ is used throughout to indicate intrinsic-mean correction.

\subsection{Fast-Folding Proteins}
With this framework established, we evaluated whether \emph{Orientation features} capture local conformational changes during protein folding. We analyzed nine fast-folding protein trajectories \cite{LindorffLarsen2011} using the Algorithm for Multiple Unknown Signals Extraction (AMUSE), which combines Principal Component Analysis (PCA) and Time-lagged Independent Component Analysis (TICA) \cite{Hyvrinen2001,PrezHernndez2013}. We compared Orientation and Orientation~$\odot$ representations against commonly used alternatives, including \Ca~coordinates, backbone torsion angles, and Riemannian Pointcloud descriptors \cite{Diepeveen2023}. Torsion angles were embedded using sine--cosine representations to account for periodicity prior to analysis.

\begin{figure}
	\centering
    \includegraphics[width=\textwidth]{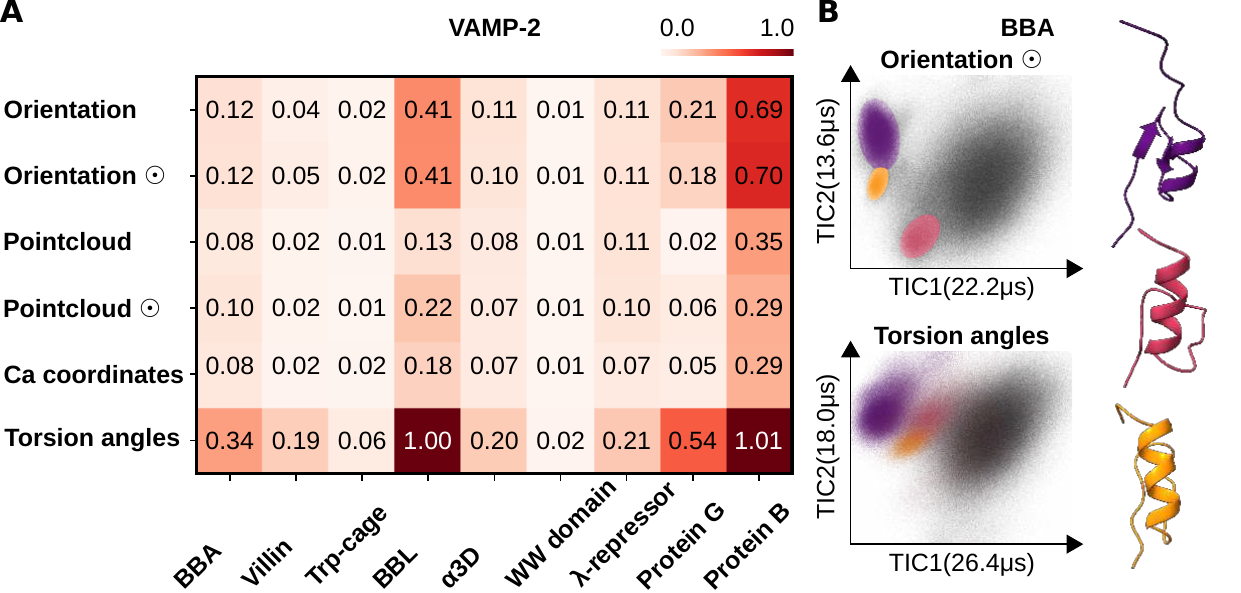}
    \caption{
        \textbf{Kinetic Characterization of Fast-Folding Proteins.}
        (\textbf{A}) VAMP-2 scores for each molecular representation (y-axis) evaluated at the comparison lag time. Systems are shown along the x-axis.
        (\textbf{B}, \textit{left}) TICA landscapes of 1FME for Orientation~$\odot$ and torsion angle representations. The x- and y-axes show TIC1 and TIC2, respectively, with implied timescales annotated. Points are colored by Orientation~$\odot$ cluster assignment; darker shades indicate unassigned frames.
        (\textbf{B}, \textit{right}) Representative centroid structures for each cluster, shown in cartoon representation.
    }
    \label{fig:figure_2}
\end{figure}

For each trajectory and representation, we performed a hyperparameter search over lag times ranging from 1\% to 10\% of the total trajectory length. Clear implied-timescale convergence was generally not observed, with the exception of the Pointcloud and \Ca~representations for $\alpha$3D. To enable a consistent comparison, we evaluated variational approach for Markov processes (VAMP-2) scores at the largest lag time considered for each system (Figure~\ref{fig:figure_2}A). Across all trajectories, torsion-angle representations achieved the highest VAMP-2 scores, indicating strong capture of slow kinetic variance. Orientation and Orientation~$\odot$ yielded intermediate scores, comparable to or slightly exceeding those obtained with Pointcloud and \Ca~representations.

Since similar implied timescales were recovered across representations, we next examined their ability to resolve distinct conformational partitioning. For each trajectory, we performed clustering in the Orientation~$\odot$ TICA space using HDBSCAN \cite{Campello2013}, followed by Gaussian mixture model expansion. Cluster assignments were then projected onto the TICA embeddings obtained from other representations. This procedure does not optimize clustering independently for each representation, but instead assess whether alternative representations capture the same or different structural variation.

Across the nine systems, three qualitative regimes emerged. In the WW domain (2F21), $\lambda$-repressor (1LMB), and protein G (1MIO), trajectories remained largely confined to the folded ensemble (Supplementary Figures~\ref{fig:figure_s1}--\ref{fig:figure_s3}). All representations identified a single dominant state and captured folding--unfolding transitions along the leading TIC, with additional minor clusters appearing in some representations but lacking clear secondary-structure signatures. In Villin (2F4K), Trp-Cage (2JOF), and BBL (2WXC), clustering results were broadly consistent across representations (Supplementary Figures~\ref{fig:figure_s4}--\ref{fig:figure_s6}). For example, Villin exhibited two coherent clusters corresponding to the canonical fold and a transient conformation, while BBL was dominated by two non-repeating unfolded conformations that produced isolated signals in all representations.

The clearest representation-dependent differences arose in BBA (1FME), $\alpha$3D (2A3D), and protein B (1PRB) (Supplementary Figures~\ref{fig:figure_s7}--\ref{fig:figure_s9}). These systems lack higher-order organization, and folding is largely governed by secondary-structure formation, making DSSP assignments a useful proxy for characterizing local conformational clusters. We emphasize that DSSP provides a reference for local structural motifs rather than a definitive ground truth for full conformational organization.

In BBA, a $\beta\beta\alpha$-fold protein \cite{Sarisky2001}, DSSP analysis identified three conformational clusters (Figure~\ref{fig:figure_2}B, Supplementary Figures~\ref{fig:figure_s7} and \ref{fig:figure_s10}). Orientation~$\odot$ resolved all three states, with the leading TIC separating folded from partially folded conformations and the second TIC distinguishing between two distinct folded states. Orientation and Pointcloud-based representations also separated these states, whereas \Ca~coordinates did not recover the two folded conformations and torsion-angle representations showed overlap between partially folded and non-canonical folded states.

$\alpha$3D, a three-helix bundle protein \cite{Walsh1999}, exhibited three DSSP-defined clusters corresponding to the canonical fold and two non-canonical conformations involving partial helix unwinding (Supplementary Figures~\ref{fig:figure_s8} and \ref{fig:figure_s11}). Orientation~$\odot$ resolved two stable states and captured a transient intermediate as a localized excursion in the leading TICs. The Orientation representation preserved this structure, whereas \Ca, torsion-angle, and Pointcloud representations did not clearly separate these states. While dynamic cross-correlation matrices showed minimal differences between clusters, cross-orientation matrices revealed localized reorientation near Pro51, highlighting sensitivity to rotational changes not evident in translational correlations.

In protein B, another three-helix bundle protein \cite{Johansson1997}, DSSP analysis identified a canonical folded state and a non-canonical conformation involving partial unfolding of the third helix (Supplementary Figures~\ref{fig:figure_s9} and \ref{fig:figure_s12}). Orientation~$\odot$ resolved these states as distinct clusters, with the leading TIC capturing the transition between them. Other representations did not distinguish these conformations. Cross-orientation analysis further revealed localized rotational differences in the Thr32--Asn37 region, whereas dynamic cross-correlation patterns differed only modestly between states.

\subsection{SARS-CoV-2 nsp13}
Having established that Orientation~$\odot$ resolves local conformational changes in small proteins, we next assessed its ability to capture coordinated domain motions in a larger, multi-domain system. We analyzed five MD trajectories of the SARS-CoV-2 nsp13 helicase \cite{Chen2022}, comparing Orientation and Orientation~$\odot$ representations against \Ca~coordinates, torsion angles, and Riemannian Pointcloud descriptors. Feature spaces were analyzed using eigendecomposition of Gram matrices and principal component analysis (PCA).

nsp13 is a superfamily 1B helicase composed of five domains: an N-terminal zinc-binding (ZB) domain, a stalk domain, a 1B domain, and two RecA-like ATPase domains (RecA1 and RecA2). Structural studies have shown that nsp13 undergoes large-scale conformational rearrangements between open and closed states, driven primarily by coordinated motions of the 1B and RecA2 domains \cite{Chen2022}. These collective, long-range motions provide a stringent test for representations intended to capture global protein dynamics.

\begin{figure}
	\centering
    \includegraphics[width=\textwidth]{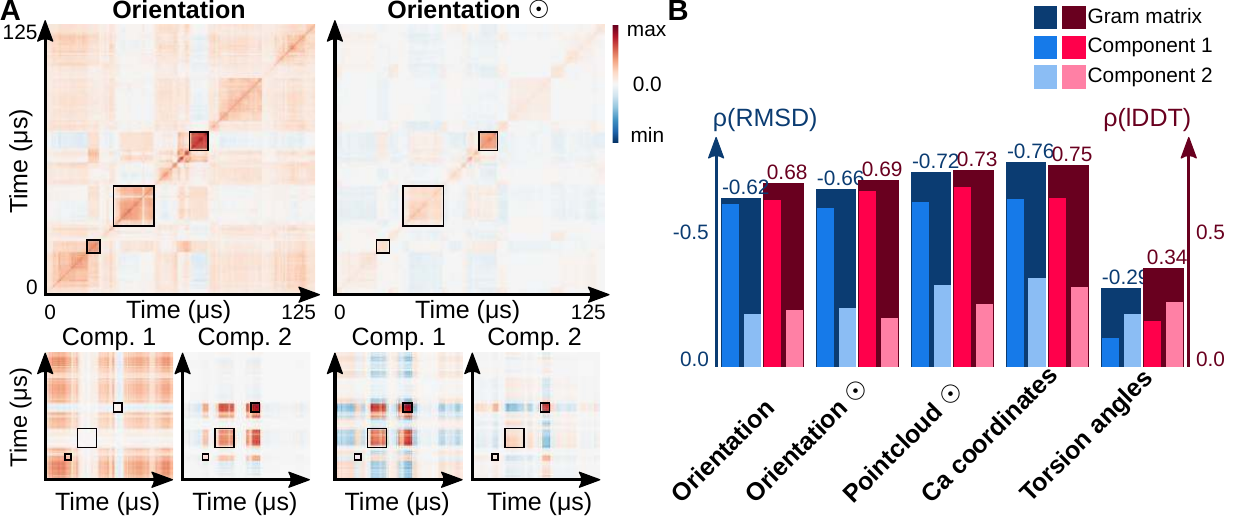}
    \caption{
        \textbf{nsp13 Gram Matrices Analysis.}
        (\textbf{A}) Gram matrices for Orientation and Orientation~$\odot$ representations. Axes represent time across merged trajectories; color intensity indicates pairwise inner product values. Black squares highlight closed-like conformations. Top row: full Gram matrices. Bottom row: first two rank-1 projection matrices.
        (\textbf{B}) Spearman correlation coefficients between Gram matrices and structural similarity measures. Blue bars: correlation with RMSD; red bars: correlation with lDDT. Inner bars show correlations for rank-1 components.
    }
    \label{fig:figure_3}
\end{figure}

We first evaluated whether similarities measured in each representation space reflect physically meaningful structural relationships. For each representation, we computed pairwise Gram matrices using the Euclidean inner product on raw feature vectors (Supplementary Figure~\ref{fig:figure_s13}) and compared them with reference matrices derived from Root Mean Square Deviation (RMSD)  and local Distance Difference Test (lDDT) scores \cite{Adhikari2021} (Supplementary Figures~\ref{fig:figure_s14} and \ref{fig:figure_s15}). To assess whether dominant modes in feature space align with structural similarity, we further correlated rank-1 projection matrices obtained from Gram matrix eigendecomposition with the reference matrices.

Three distinct similarity profiles emerged. Orientation~$\odot$, Pointcloud, and \Ca~coordinates produced closely related Gram matrices and exhibited moderate to strong correlations with both RMSD and lDDT ($|\rho| = 0.66$--$0.76$). These representations yielded balanced similarity patterns containing multiple high-similarity clusters corresponding to closed-like conformations. Within this group, Orientation~$\odot$ displayed greater heterogeneity, indicating sensitivity to finer structural distinctions. The standard Orientation representation formed a second profile, characterized by uniformly higher similarity values that compressed distinctions among closed states. Torsion-angle representations formed a third profile, exhibiting weak correlations with RMSD ($\rho = -0.29$) and lDDT ($\rho = 0.34$) and producing similarity patterns poorly aligned with structural rankings.

\begin{figure}
	\centering
    \includegraphics[width=\textwidth]{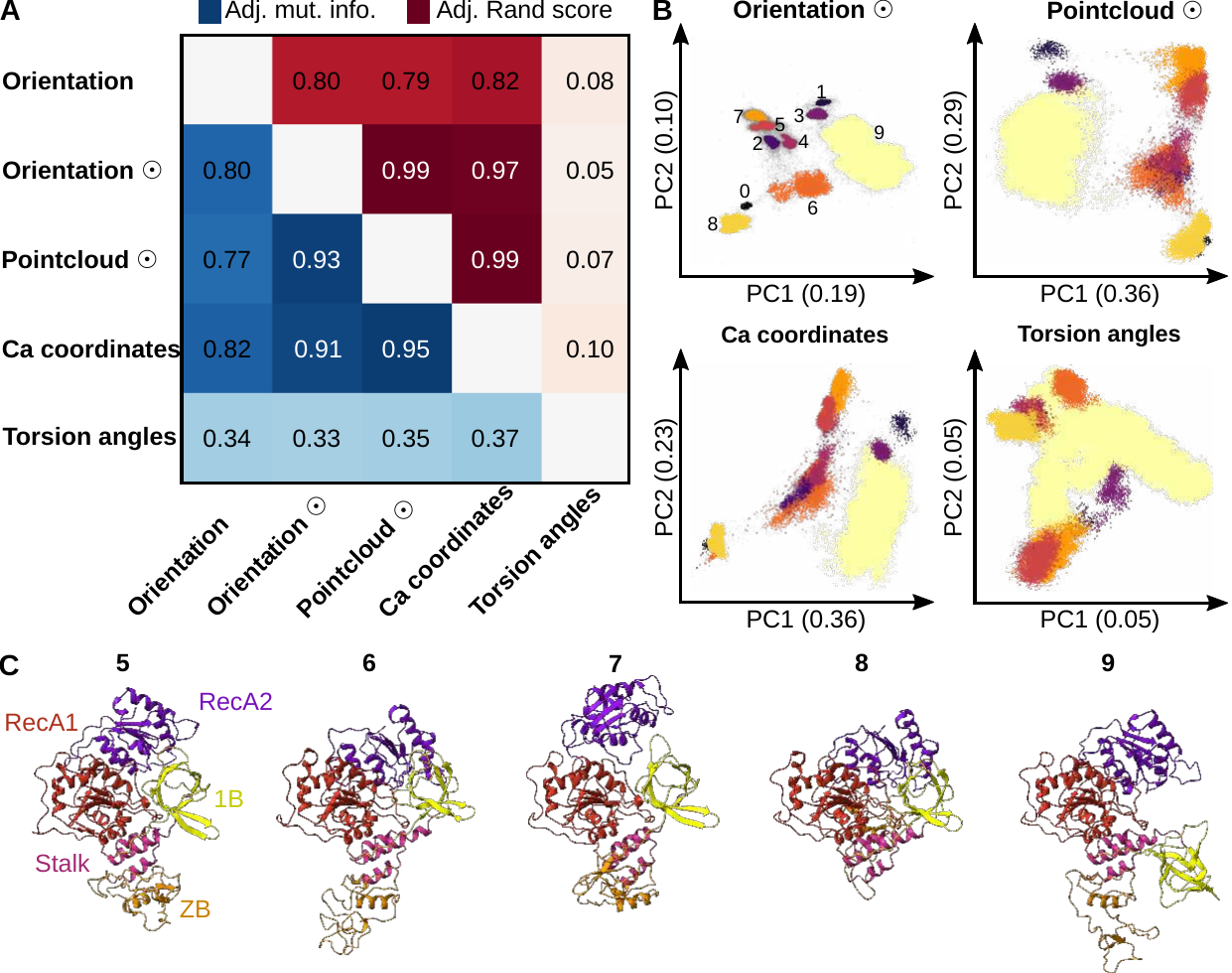}
    \caption{
        \textbf{nsp13 Clusters.}
        (\textbf{A}) Clustering agreement across representations. Upper triangle: Adjusted Rand Score; lower triangle: Adjusted Mutual Information.
        (\textbf{B}) PC1–PC2 projections for each representation. Points are colored by Orientation~$\odot$ cluster assignment.
        (\textbf{C}) Centroid structures of the five most populated Orientation~$\odot$ clusters in panel B, aligned on the stalk and RecA1 domains. The monomeric nsp13 protein is shown in a ribbon representation, with domains color-coded as follows: ZB (orange), Stalk (pink), 1B (yellow), RecA1 (red) and RecA2 (Purple).
    }
    \label{fig:figure_4}
\end{figure}

We next characterized the conformational landscape using PCA and clustering. Since the Gram matrix defines a linear kernel, decomposition of the Gram matrix corresponds to kernel PCA and enables direct interpretation of dominant modes of variability. Clustering agreement across representations was quantified using Adjusted Mutual Information (AMI) and Adjusted Rand Score (ARS) \cite{Vinh2009,Hubert1985,scikit-learn} (Figure~\ref{fig:figure_4}A). Orientation~$\odot$, Pointcloud, and \Ca~coordinates showed strong mutual agreement, whereas the standard Orientation representation exhibited intermediate consistency and torsion angles showed limited agreement. Silhouette scores computed using RMSD as the distance metric were low across all representations, indicating that clustering partitions only weakly reflect RMSD-defined similarity. Orientation~$\odot$, Pointcloud, and \Ca~coordinates achieved the highest silhouette scores (0.14--0.15), Orientation scored near zero (0.01), and torsion angles yielded a negative value (-0.13).

Despite these overall similarities, Orientation~$\odot$ provided complementary discriminative information. Differences relative to Pointcloud and \Ca~coordinates were most pronounced in clusters 2, 4, and 6, where the latter representations tended to merge structurally distinct states (Figure~\ref{fig:figure_4}B, Supplementary Figure~\ref{fig:figure_s16}). Notably, torsion-angle representations aligned with Orientation~$\odot$ assignments for these clusters, suggesting that torsional and orientation-based features can emphasize overlapping but distinct aspects of the underlying conformational transitions.

Structural inspection of Orientation~$\odot$ cluster centroids revealed that four of the five most populated clusters corresponded to closed-like conformations, while one represented an open-like state (Figure~\ref{fig:figure_4}C, Supplementary Figure~\ref{fig:figure_s17}). Conformational variability was concentrated in the ZB, 1B, and RecA2 domains, with the ZB exhibiting the largest amplitude of motion. Reorientations of the 1B and RecA2 domains were captured predominantly by the first two principal components, with transitions among centroids consistent with previously described open--closed rearrangements. These results indicate that Orientation~$\odot$ captures coordinated domain motions in nsp13 while providing sensitivity to structural distinctions that are partially obscured in translational representations.

\subsection{Protein--Protein Association}
The preceding analyses focused on internal protein dynamics. We next examined a more challenging regime: protein--protein association, where internal conformational changes are minimal and the dominant signal arises from subtle interface rearrangements \cite{Pan2019}. This setting provides a stringent test for representation sensitivity under conditions of limited structural variability.

\begin{figure}
	\centering
    \includegraphics[width=\textwidth]{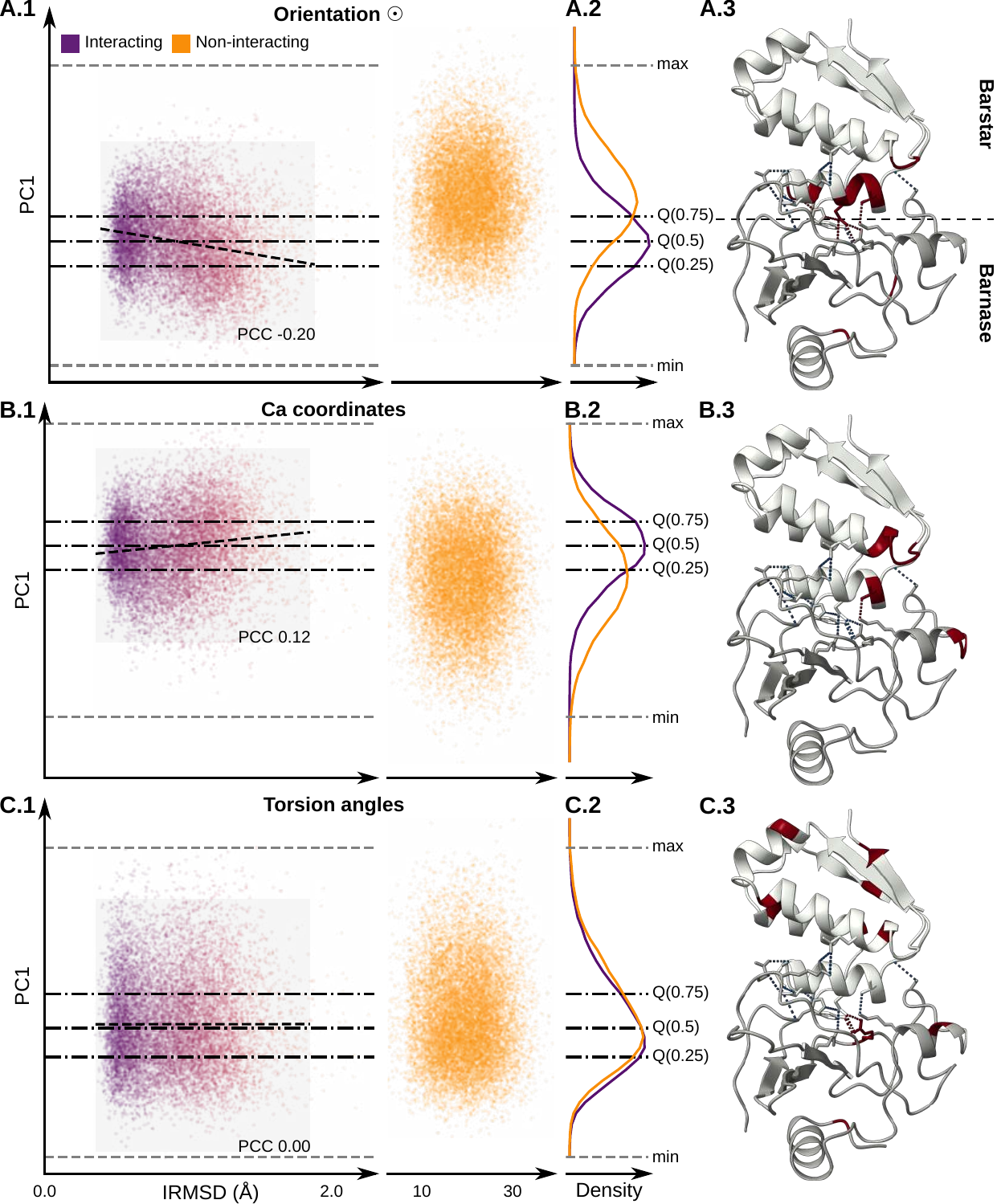}
    \caption{
        \textbf{Molecular Representations for Barnase–Barstar Association.}
        Results for (\textbf{A}) Orientation~$\odot$, (\textbf{B}) C$\alpha$ coordinates, and (\textbf{C}) torsion angles. In all panels:
        (\textit{.1}) PC1 (y-axis) versus IRMSD (x-axis). Interacting frames shown in purple-to-orange gradient (by IRMSD); non-interacting frames in orange. Gray region: interquartile range $[Q_1{-}1.5\,\mathrm{IQR},\, Q_3{+}1.5\,\mathrm{IQR}]$; Pearson Correlation Coefficient (PCC) annotated.
        In all panels: (\textit{.2}) PC1 density distributions for interacting/non-interacting states with quartile lines.
        In all panels: (\textit{.3}) Reference structure (Barstar: light gray; Barnase: dark gray). Top 10 PC1-contributing residues highlighted in red; hydrogen bonds shown as dashed lines (red: involving contributors; blue: other).
    }
    \label{fig:figure_5}
\end{figure}

We analyzed four sets of association trajectories involving compact, structurally stable proteins: the Barnase--Barstar dimer, two insulin dimer trajectories, and the RNaseHI--SSB complex \cite{Pan2019}. To assess the contribution of orientation-based descriptors, we decomposed \emph{Orientation features} into rotation axis ($u$) and rotation angle ($\theta$) components, exploiting the axis--angle structure of the logarithmic map. We applied a two-step principal component analysis (PCA), first performing PCA independently on each protein and then combining the resulting components for a joint PCA of the complex.

To quantify how effectively each representation captured association dynamics, we defined bound and unbound states using system-specific interface RMSD (IRMSD) thresholds and computed the mutual information (MI) between principal component projections and state labels (Supplementary Figure~\ref{fig:figure_s18}). Across all systems and representations, MI computed using the leading principal component (PC1) was indistinguishable from that obtained using PC1 and PC2 jointly, indicating that the association signal was effectively one-dimensional.

Performance varied across systems and representations, reflecting differences in the dominant association mechanisms. Orientation~$\odot$ ($u$) achieved the highest MI for the Barnase--Barstar system and for one insulin dimer trajectory. In contrast, Pointcloud and \Ca~coordinate representations performed best for the second insulin dimer trajectory, while torsion angles yielded the strongest separation for the RNaseHI--SSB complex.

Analysis of PC1 distributions further illustrated these trends. In Barnase--Barstar (Supplementary Figure~\ref{fig:figure_s19}), Orientation~$\odot$ ($u$) provided the clearest separation between bound and unbound states, although partial overlap reduced overall MI. In the first insulin dimer trajectory set (Supplementary Figure~\ref{fig:figure_s20}), all representations separated bound and unbound states, with Orientation~$\odot$ ($u$) yielding the largest margin of separation. In the second insulin dimer trajectory set (Supplementary Figure~\ref{fig:figure_s21}), a non-canonical bound state with elevated IRMSD emerged. Pointcloud and \Ca~coordinates separated this state from others but did not distinguish canonical bound from unbound conformations, whereas Orientation~$\odot$ ($u$) differentiated all three states, albeit with some overlap.

In the RNaseHI--SSB system (Supplementary Figure~\ref{fig:figure_s22}), torsion-angle representations most effectively separated bound and unbound states, while other representations exhibited substantial overlap. This behavior is consistent with a folding-upon-binding mechanism in which local backbone rearrangements dominate the association signal.

Across all structurally stable complexes examined here, the rotation axis component ($u$) carried the dominant discriminative information, whereas the rotation angle ($\theta$) contributed minimally. This observation is specific to systems with limited internal motion and is not expected to generalize to association processes involving larger conformational changes.

To examine the structural basis of Orientation~$\odot$ performance, we focused on the Barnase--Barstar system (Figure~\ref{fig:figure_5}). Correlation analysis between PC1 projections and IRMSD for interacting frames revealed a moderate correlation for Orientation~$\odot$, which was substantially reduced for \Ca~coordinates and absent for torsion-angle representations. Residue-level contributions to PC1 further distinguished the representations: for Orientation~$\odot$, the most influential residues localized to the Barstar interface and participated in multiple hydrogen bonds, whereas residues identified by \Ca~coordinates and torsion angles were more spatially dispersed and did not preferentially localize to the interface.

\begin{figure}
	\centering
    \includegraphics[width=\textwidth]{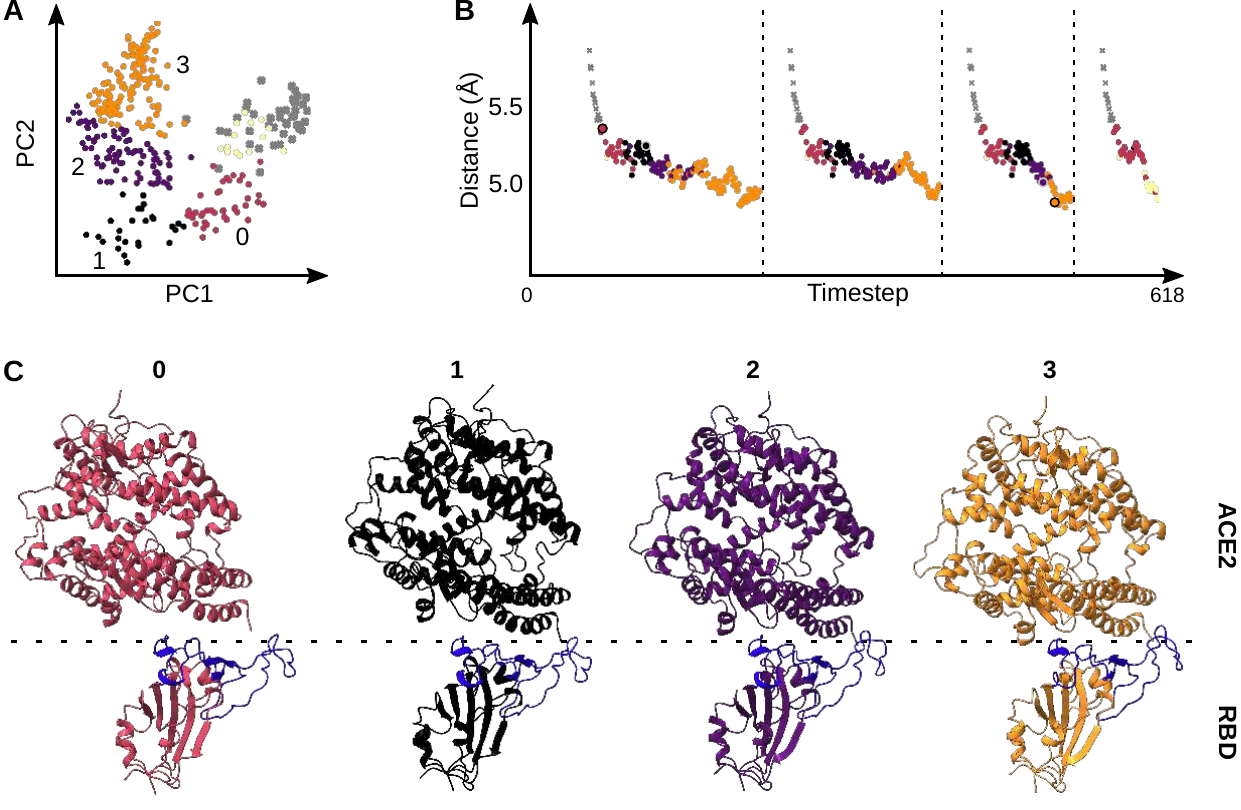}
    \caption{
        \textbf{Landscape of RBD-ACE2 Association.}
        (\textbf{A}) PC1–PC2 projection of the Orientation~$\odot$ (u) representation. Points colored by cluster assignment.
        (\textbf{B}) Time series of the distance between protein centers of geometry. The x-axis represents time; y-axis shows distance. Colors indicate cluster assignment.
        (\textbf{C}) Centroid structures of the four bound-state clusters. Upper region: ACE2 receptor; lower region: RBD. The receptor-binding motif (RBM) is highlighted in blue.
    }
    \label{fig:figure_6}
\end{figure}

Finally, we extended this analysis to the RBD--ACE2 association using Orientation and Orientation~$\odot$ ($u$) representations across four trajectories (Supplementary Figure~\ref{fig:figure_s23}) \cite{deshawcovid}. These systems exhibit minimal internal motion, with association characterized by a subtle conformational change in the receptor-binding motif. Using the same two-step PCA framework, PC1 consistently captured the binding process, while PC2 reflected a slower relaxation motion. Both Orientation representations produced similar clustering behavior and correctly identified the terminal frames of the final trajectory as unbound.

\subsection{ManiProt Library}
To facilitate broader use of the proposed representations, we developed ManiProt, a computational library implementing the methods described above. ManiProt is written in Python and leverages NumPy \cite{numpy} and Numba \cite{numba} to achieve efficient numerical performance.

\begin{figure}
	\centering
    \includegraphics[width=\textwidth]{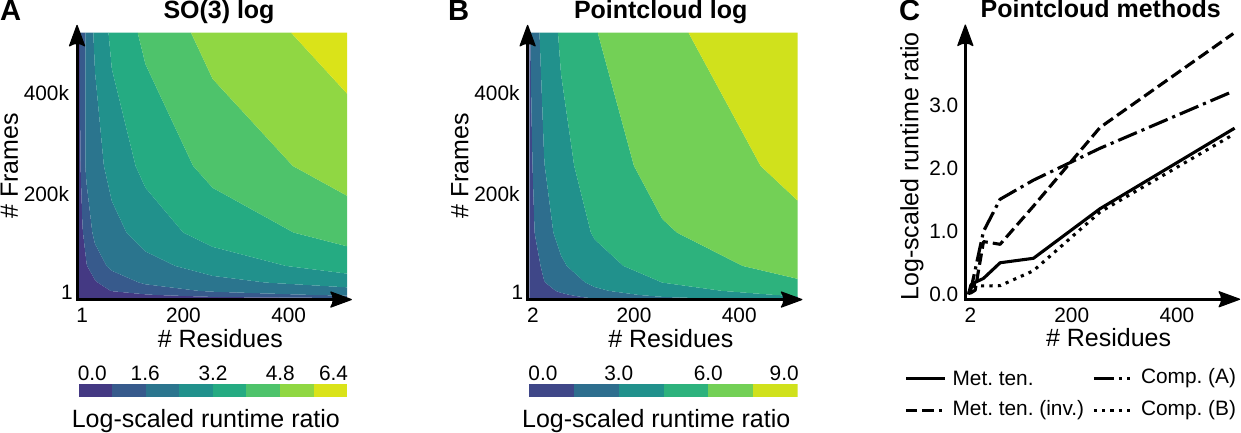}
    \caption{
        \textbf{ManiProt Profile.}
        (\textbf{A, B}) Computational scaling of the SO(3) (\textbf{A}) and Pointcloud (\textbf{B}) logarithmic maps. The x-axis shows the number of residues; y-axis shows the number of frames. Color indicates the log-ratio of execution time relative to the smallest problem size (1 (SO(3)) and 2 (Pointcloud) residue, 1 frame).
        (\textbf{C}) Scaling of Pointcloud-associated operations (metric tensor (Met. ten.) computation and inversion (inv.)). The x-axis shows the number of residues; y-axis shows the log-ratio of execution time relative to 4 residues.
    }
    \label{fig:figure_7}
\end{figure}

We characterized computational scaling with a focus on the logarithmic map, which constitutes the dominant cost in projecting structures onto the tangent space. For \emph{Orientation features}, runtime scaled linearly with both the number of residues and the number of trajectory frames. In contrast, Pointcloud representations exhibited quadratic scaling with residue count, reflecting the $O(r^2)$ cost of pairwise distance evaluations, while remaining linear in the number of frames.

Parallelization was applied over trajectory frames rather than residues, which limits efficiency gains for very large systems. For Pointcloud-based representations, additional computational overhead arose from construction of the inverse metric tensor, dominated by eigendecomposition. These scaling characteristics delineate the practical regimes in which each representation is computationally advantageous. 

\section{Discussion}

Molecular Dynamics simulations generate high-dimensional trajectories that capture protein motion at atomic resolution, providing critical insights into biomolecular function. Extracting meaningful information from these trajectories, however, remains challenging, since subtle conformational changes may be obscured by noise, global rotations, or high-dimensional correlations. Standard representations, such as Cartesian coordinates, torsion angles, or Pointclouds, emphasize different aspects of motion and can yield divergent interpretations. These challenges motivated the development of \emph{Orientation features}, a rotation-aware, geometrically grounded representation that encodes residue-level backbone rotations on the product manifold $\mathrm{SO}(3)^n$ and projects them onto a tangent space.

\emph{Orientation features} provides a compact, rotation-focused encoding of protein motion that is compatible with conventional machine learning methods. Across three complementary regimes, i.e., fast-folding proteins, a multi-domain helicase, and protein--protein association, it captures aspects of dynamics that complement existing representations. In fast-folding proteins, Orientation~$\odot$ resolved structural partitions that were indistinguishable with other features under the same clustering protocol. The TIC decomposition yielded interpretable axes corresponding to secondary-structure transitions, highlighting the ability of rotation-focused features to reveal local conformational variability. The superior VAMP-2 performance of torsion angles reflects their direct correspondence to the dihedral transitions governing folding. The comparable VAMP-2 scores of Orientation, Pointcloud and \Ca~representations suggest these features operate on similar kinetic timescales while emphasizing different structural aspects.

For the multi-domain nsp13 helicase, Orientation~$\odot$ captured coordinated rearrangements of the ZB, 1B, and RecA2 domains. PCA and clustering revealed clusters that \Ca~coordinates and Pointcloud features merged, demonstrating sensitivity to subtle conformational differences. Although these putative intermediate states were not validated as metastable, the results indicate that rotational features can highlight intermediate motions that may be overlooked in conventional representations.

In protein--protein association, \emph{Orientation features} detected conformational changes despite minimal internal motion. The rotation axis component ($u$) carried the primary discriminative signal, whereas the rotation angle ($\theta$) contributed minimally. Residue-level contributions to principal components localized to the binding interface, showing that \emph{Orientation features} can identify atoms involved in molecular recognition. Performance was system-dependent, consistent with the principle that different representations capture distinct aspects of motion.

We developed the ManiProt library which provides an efficient implementation of these methods, with \emph{Orientation features} scaling linearly in both residue count and trajectory length. In contrast, Pointcloud representations scale quadratically with residue count, limiting their practical application to large proteins and long trajectories.

A key methodological insight from our study is that no single representation is universally optimal. Different features emphasize distinct dynamical mechanisms, i.e., torsion angles for local backbone transitions, \Ca~coordinates or Pointclouds for global shape changes, and \emph{Orientation features} for coordinated rotational or domain-level motions. Comparative analysis across multiple representations therefore provides complementary insights and reduces the risk of representation-specific artifacts.

Several limitations should be noted. The product structure $\mathrm{SO}(3)^n$ imposes a decoupled metric that neglects direct geometric coupling between residues. Energy-related features, such as distances or torsions, are not explicitly encoded. Those limitations are shared by other common representations, i.e., torsion angles treat each dihedral independently, and \Ca~coordinates do not encode covalent connectivity. Solving them would require constructing a more elaborate manifold, substantially increasing the mathematical and computational complexity. Even though our analysis focuses on linear projections (PCA, AMUSE) for interpretability, \emph{Orientation features} is compatible with nonlinear or deep learning models, including kernel methods, autoencoders, and graph neural networks. Nonlinear approaches such as VAMPnets, time-lagged autoencoders, or diffusion maps could be applied to \emph{Orientation features} and may yield further improvements, but would confound the comparison by introducing method-specific inductive biases.

In summary, \emph{Orientation features} provides a geometrically principled, rotation-aware framework for analyzing MD trajectories. By capturing residue-level rotations in a compact representation and evaluating them alongside conventional descriptors, this approach enables systematic, interpretable insights into both local and global protein dynamics across diverse systems.

\section*{Materials and Methods}

\subsection*{Mathematical Formulation}
\paragraph{Local Coordinate System}
Given a molecular dynamics trajectory with frames $t = 1, \ldots, T$, let $\mathrm{C}_{(r,t)}$, $\mathrm{C\alpha}_{(r,t)}$, and $\mathrm{N}_{(r,t)}$ denote the coordinates of the carbonyl carbon ($\mathrm{C}$), alpha carbon ($\mathrm{C\alpha}$), and amide nitrogen ($\mathrm{N}$) atoms of residue $r$ at frame $t$, respectively.

We construct a Local Coordinate System for each residue as follows \cite{Ingraham2019,Jumper2021}. First, define the unit vectors $\mathbf{u}_{(r,t)}$ and $\mathbf{w}_{(r,t)}$ pointing from $\mathrm{C\alpha}$ to $\mathrm{N}$ and from $\mathrm{C\alpha}$ to $\mathrm{C}$, respectively:
\begin{equation}
    \mathbf{u}_{(r,t)} = \frac{\mathrm{N}_{(r,t)} - \mathrm{C\alpha}_{(r,t)}}{\left\|\mathrm{N}_{(r,t)} - \mathrm{C\alpha}_{(r,t)}\right\|}, \quad
    \mathbf{w}_{(r,t)} = \frac{\mathrm{C}_{(r,t)} - \mathrm{C\alpha}_{(r,t)}}{\left\|\mathrm{C}_{(r,t)} - \mathrm{C\alpha}_{(r,t)}\right\|}.
\end{equation}
Next, compute the normal vector $\mathbf{n}_{(r,t)} = \mathbf{u}_{(r,t)} \times \mathbf{w}_{(r,t)} / \left\|\mathbf{u}_{(r,t)} \times \mathbf{w}_{(r,t)}\right\|$, orthogonal to the plane spanned by $\mathbf{u}_{(r,t)}$ and $\mathbf{w}_{(r,t)}$. Finally, define $\mathbf{v}_{(r,t)} = \mathbf{u}_{(r,t)} \times \mathbf{n}_{(r,t)}$ to complete the orthonormal basis. These three vectors form the LCS matrix:
\begin{equation}
    \mathrm{LCS}_{(r,t)} = \begin{bmatrix} \mathbf{u}_{(r,t)} & \mathbf{n}_{(r,t)} & \mathbf{v}_{(r,t)} \end{bmatrix},
\end{equation}
which is an element of the special orthogonal group $\mathrm{SO}(3)$, representing the orientation of residue $r$ at frame $t$.

\paragraph{Logarithmic Map}
The logarithmic map $\log: \mathrm{SO}(3) \to \mathfrak{so}(3) \cong \mathbb{R}^3$ projects a rotation matrix onto its axis-angle representation in the Lie algebra. For a rotation matrix $R \in \mathrm{SO}(3)$, the rotation angle $\theta \in [0, \pi]$ is computed from the trace:
\begin{equation}
    \cos \theta = \frac{1}{2} (\mathrm{tr}(R) - 1), \quad
    \sin \theta = \frac{1}{2} \sqrt{(3 - \mathrm{tr}(R))(1 + \mathrm{tr}(R))},
\end{equation}
with $\theta = \mathrm{atan2}(\sin \theta, \cos \theta)$. The latter formulation for $\sin\theta$ avoids numerical instabilities associated with $\arccos$ near $\theta = 0$ and $\theta = \pi$.

For a skew-symmetric matrix $\Omega \in \mathfrak{so}(3)$, the vee operator $(\cdot)^\vee$ extracts the corresponding axial vector:
\begin{equation}
    \Omega = \begin{bmatrix}
        0      & -\omega_z & \omega_y \\
        \omega_z    & 0    & -\omega_x \\
        -\omega_y   & \omega_x  & 0
    \end{bmatrix}
    \quad \Longrightarrow \quad
    \Omega^\vee = \begin{bmatrix}
        \omega_x \\
        \omega_y \\
        \omega_z
    \end{bmatrix}.
\end{equation}

The logarithmic map $\boldsymbol{\omega} = \log(R)^\vee \in \mathbb{R}^3$ is defined piecewise to ensure numerical stability:
\begin{equation}
    \boldsymbol{\omega} =
    \begin{cases}
    \mathbf{0}, & \text{if } \theta < \varepsilon, \\[4pt]
    \dfrac{\theta}{2 \sin \theta} [R - R^\top]^\vee, & \text{if } \varepsilon \leq \theta \leq \pi - \varepsilon, \\[8pt]
    \theta \cdot \mathbf{n}, & \text{if } \theta > \pi - \varepsilon,
    \end{cases}
\end{equation}
where $\varepsilon$ is a small threshold. In the first case ($\theta \approx 0$), the rotation is negligible. In the second case, the standard formula applies. In the third case ($\theta \approx \pi$), the axis $\mathbf{n}$ is extracted from the symmetric part $S = R + R^\top + (1 - \mathrm{tr}(R)) I$, with components $n_i = \sqrt{S_{ii} / (3 - \mathrm{tr}(R))}$; the sign of $\mathbf{n}$ is determined by consistency with $[R - R^\top]^\vee$ \cite{Blanco2021,Nurlanov2021}.

The logarithmic map is well-defined and smooth for rotation angles strictly less than $\pi$. At $\theta = \pi$, the map reaches the cut locus of the identity, where the rotation axis becomes ambiguous (antipodal points on SO(3) represent the same rotation). For the rare cases where residue orientations approach this limit, the piecewise formulation provides a numerically stable approximation, with negligible impact on downstream analyses.

\paragraph{Intrinsic Mean Computation}
The intrinsic mean on \SO{3} is the rotation matrix that minimizes the sum of squared geodesic distances to all observations. For a set of rotation matrices $\{R_1, \ldots, R_T\}$, the intrinsic mean $\bar{R}$ is defined as:
\begin{equation}
    \bar{R} = \arg\min_{R \in \mathrm{SO}(3)} \sum_{t=1}^{T} d_{\mathrm{geo}}(R, R_t)^2,
\end{equation}
where $d_{\mathrm{geo}}(R, R_t) = \|\log(R^\top R_t)\|$ is the geodesic distance on \SO{3}. Since no closed-form solution exists, we compute the intrinsic mean iteratively using Riemannian gradient descent \cite{Fletcher2004}. At each iteration $k$, the current estimate $\bar{R}^{(k)}$ is updated by:
\begin{equation}
    \bar{R}^{(k+1)} = \exp_{\bar{R}^{(k)}}\left(\eta \cdot \mathbf{g}^{(k)}\right), \quad \text{where} \quad \mathbf{g}^{(k)} = \frac{1}{T} \sum_{t=1}^{T} \log_{\bar{R}^{(k)}}(R_t).   
\end{equation}
Here, $\log_{\bar{R}}(R_t) = \log(\bar{R}^\top R_t)$ computes the tangent vector from $\bar{R}$ to $R_t$, $\mathbf{g}^{(k)} \in \mathbb{R}^3$ is the Riemannian gradient (the average tangent direction to all observations), $\eta$ is the learning rate, and $\exp_{\bar{R}}(\mathbf{v}) = \bar{R} \cdot \exp(\mathbf{v}^\wedge)$ maps the tangent vector back to \SO{3} via left translation. The algorithm terminates when the gradient norm $\|\mathbf{g}^{(k)}\|$ falls below a convergence threshold or the maximum number of iterations is reached. For Orientation~$\odot$ features, this procedure is applied independently to each residue, yielding a per-residue reference frame $\bar{R}_r$ that represents the average orientation across the trajectory.

\subsection*{Data}

\paragraph{Fast-Folding Proteins}
We analyzed nine fast-folding protein trajectories generated by D.E. Shaw Research (Supplementary Table \ref{tab:table_s1}) \cite{LindorffLarsen2011}. The dataset comprises simulations of small proteins ranging from 125 to 1154 $\mu\mathrm{s}$: BBA (1FME, 325~$\mu\mathrm{s}$), Villin (2F4K, 125~$\mu\mathrm{s}$), Trp-cage (2JOF, 208~$\mu\mathrm{s}$), BBL (2WXC, 429~$\mu\mathrm{s}$), $\alpha$3D (2A3D, 707~$\mu\mathrm{s}$), WW domain (2F21, 1137~$\mu\mathrm{s}$), $\lambda$-repressor (1LMB, 643~$\mu\mathrm{s}$), Protein G (1MIO, 1154~$\mu\mathrm{s}$) and Protein B (1PRB, 104~$\mu\mathrm{s}$).

\paragraph{SARS-CoV-2 nsp13}
We analyzed five trajectories of the SARS-CoV-2 nsp13 helicase generated by D.E. Shaw Research (Supplementary Table \ref{tab:table_s1}) \cite{Chen2022,deshawcovid}. The dataset comprises publicly available simulations of 25~$\mu\mathrm{s}$ (accession codes: 12212701, 12212688, 12212691, 12212689, and 1221269).

\paragraph{Protein-Protein Association}
We first analyzed protein-protein association trajectories from dimmer simulations generated by D.E. Shaw Research (Supplementary Table \ref{tab:table_s1}) \cite{Pan2019}. The dataset comprises four systems and each system had three associated trajectories ranging from 140.7~$\mu\mathrm{s}$ to 484.2~$\mu\mathrm{s}$: Barnase-Barstar (3$\times$400.0~$\mu\mathrm{s}$), Insulin dimer (0) (3$\times$140.7~$\mu\mathrm{s}$), Insulin dimer (1) (3$\times$484.2~$\mu\mathrm{s}$) and RNaseHI-SSB complex (3$\times$385.6~$\mu\mathrm{s}$).
We then analyzed four representative simulations of the SARS-CoV-2 receptor-binding domain bound to the ACE2 receptor (Supplementary Table \ref{tab:table_s1}) \cite{deshawcovid}. The dataset comprise publicly available data (accession code: 10895671).

\subsection*{Features}

\paragraph{Preprocessing}
Prior to computing features, all trajectories were aligned to a reference structure to remove global translational and rotational degrees of freedom. For fast-folding proteins, alignment was performed against the corresponding PDB structure. For protein-protein association systems, monomeric structures were first extracted from each trajectory and independently aligned to the corresponding monomer in the bound-state reference structure.

\paragraph{Orientation Features}
Local Coordinate Systems were computed for each residue using the backbone \N, \Ca, and \C\ atom coordinates, as described in Mathematical Formulation. To ensure rotational invariance, each LCS was expressed relative to a reference orientation by computing the relative rotation $R_{\mathrm{rel}} = R_{\mathrm{ref}}^{-1} R_{(r,t)}$, where $R_{\mathrm{ref}}$ is the reference LCS and $R_{(r,t)}$ is the LCS of residue $r$ at frame $t$. This relative rotation was then projected onto the tangent space via the logarithmic map, yielding a 3-dimensional vector per residue.

We considered two choices for the reference orientation:
\begin{itemize}
    \item \textbf{Orientation}: The reference LCS is taken from the aligned reference structure used in preprocessing.
    \item \textbf{Orientation~$\boldsymbol{\odot}$}: The reference LCS is the intrinsic mean of all LCS observations for that residue across the trajectory. For intrinsic mean computation, we applied a Riemannian gradient descent algorithm with a maximum of 256 iterations, a learning rate of 0.1, and a convergence threshold of $10^{-3}$.
\end{itemize}

Since the logarithmic map inherently centers the data at the identity element of SO(3), no additional demeaning was applied to Orientation or Orientation~$\odot$ features. Normalization was also omitted to preserve the geometric interpretation of the rotation magnitude.

For protein-protein association analyses, we additionally decomposed the Orientation~$\odot$ features into:
\begin{itemize}
    \item \textbf{Orientation~$\boldsymbol{\odot}$~($\mathbf{u}$)}: The unit direction vector (rotation axis), obtained by normalizing each 3-dimensional rotation vector to unit length.
    \item \textbf{Orientation~$\boldsymbol{\odot}$~($\boldsymbol{\theta}$)}: The rotation magnitude (angle), computed as the Euclidean norm of each rotation vector.
\end{itemize}

\paragraph{\Ca\ Coordinates}
The \Ca\ coordinates were extracted as the three-dimensional position of each \Ca\ atom after alignment. Demeaning was applied across residues.

\paragraph{Torsion Angles}
Backbone dihedral angles ($\phi$, $\psi$) were computed for each residue using MDTraj~\cite{mdtraj}. Angles were embedded using sine-cosine encoding: $(\phi, \psi) \mapsto (\sin\phi, \cos\phi, \sin\psi, \cos\psi)$, yielding a 4-dimensional feature vector per residue. Demeaning was applied to the embedded features.

\paragraph{Pointcloud Features}
Pointcloud \cite{Diepeveen2023} features represent the spatial distribution of \Ca\ atoms as a collective coordinate. For a protein with $N$ residues, the Pointcloud at frame $t$ is the $N \times 3$ matrix of \Ca\ positions. To compare Pointclouds across frames, we compute the tangent projection via the logarithmic map yielding a 3-dimensional vector per residue. We used a $\delta$-factor of 0.1.

We considered two choices for the reference configuration:
\begin{itemize}
    \item \textbf{Pointcloud}: The reference is the \Ca\ configuration from the aligned reference structure.
    \item \textbf{Pointcloud~$\boldsymbol{\odot}$}: The reference LCS is the intrinsic mean of all Pointcloud observations. For intrinsic mean computation we applied a Riemannian gradient descent algorithm with a maximum of 256 iterations, a learning rate of 1 and a convergence threshold of $1$.
\end{itemize}

As with \emph{Orientation features}, demeaning and normalization were not applied, since the displacement representation inherently centers the data at the reference configuration.

\subsection*{Fast-Folding Proteins}

For the fast-folding protein analysis, we applied a dimensionality reduction and clustering pipeline to identify metastable conformational states and characterize their structural properties. The workflow consists of four stages: (1) linear dimensionality reduction via PCA and Time-lagged independent component analysis (TICA), (2) secondary structure assignment via DSSP, (3) density-based clustering with Gaussian refinement, and (4) per-cluster correlation analysis.

\paragraph{AMUSE Algorithm}
Dimensionality reduction was performed using the AMUSE framework, which combines PCA with TICA \cite{Hyvrinen2001}.

\textit{Principal Component Analysis.} Each feature representation was first projected onto its principal components. PCA was performed with whitening enabled and a cumulative explained variance threshold of 95\%, retaining only those components necessary to capture this fraction of the total variance. Whitening decorrelates the components and normalizes their variances, ensuring that subsequent TICA analysis operates on an orthonormal basis.

\textit{Time-lagged Independent Component Analysis.} TICA identifies linear combinations of features that decorrelate most slowly over time, thereby capturing the slowest dynamical modes of the system \cite{deeptime}. Given a time series of whitened principal components $\mathbf{x}(t)$, TICA solves the generalized eigenvalue problem:
\begin{equation}
    C(\tau) \mathbf{v}_i = \lambda_i C(0) \mathbf{v}_i,
\end{equation}
where $C(0) = \langle \mathbf{x}(t) \mathbf{x}(t)^\top \rangle$ is the instantaneous covariance matrix, $C(\tau) = \langle \mathbf{x}(t) \mathbf{x}(t+\tau)^\top \rangle$ is the time-lagged covariance matrix at lag time $\tau$, and $\lambda_i$ are the eigenvalues ordered by magnitude. The implied relaxation timescale associated with each TIC component is:
\begin{equation}
    t_i = -\frac{\tau}{\ln \lambda_i}.
\end{equation}

\textit{Lag time selection.} The choice of lag time $\tau$ affects the estimated timescales: too short a lag time yields timescales that have not yet converged, while excessively long lag times reduce statistical power. We performed a parameter search over lag times ranging from 1\% to 10\% of the trajectory length. For each lag time, we computed the implied timescales and identified the plateau onset, defined as the first lag time at which the relative change in the slowest timescale remained below 10\% for at least three consecutive lag time increments. The final TICA model was fitted at the maximum plateau lag time across all representations, ensuring consistent comparison.

\textit{VAMP-2 score.} To quantify the kinetic information captured by each representation, we computed the Variational Approach for Markov Processes (VAMP-2) score~\cite{PrezHernndez2013,Wu2019,deeptime}. For a TICA model with singular values $\{\sigma_i\}$, the VAMP-2 score is defined as:
\begin{equation}
    \text{VAMP-2} = \sum_{i=1}^{k} \sigma_i^2,
\end{equation}
where $k$ is the number of retained TIC components. The VAMP-2 score provides a variational bound on the kinetic variance captured by the projection: higher scores indicate that the representation better preserves the slow dynamical modes of the system.

\paragraph{Define Secondary Structure of Proteins (DSSP)}
Secondary structure assignments were computed using the DSSP algorithm~\cite{Kabsch1983} as implemented in MDTraj~\cite{mdtraj}. DSSP classifies each residue into one of eight categories based on hydrogen bonding patterns. For visualization and analysis, we simplified the assignments to three classes: $\alpha$-helix (H), $\beta$-strand (E), and coil/other (all remaining categories).

\paragraph{Clustering}
Conformational states were identified using a two-stage clustering approach combining density-based clustering \cite{Campello2013} with Gaussian Mixture Model (GMM) \cite{scikit-learn} refinement. Clustering was performed on the Orientation~$\odot$ representation.

\textit{HDBSCAN.} Initial cluster assignment was performed using HDBSCAN~\cite{Campello2013,scikit-learn} on the two-dimensional TIC projection. The minimum cluster size was set to 0.1\% of the total number of frames. The maximum cluster size was set to 33\% by default, but increased to 90\% for trajectories dominated by the folded state (2F21, $\lambda$-repressor, and NuG2). HDBSCAN automatically determines the number of clusters and identifies outlier points that do not belong to any dense region.

\textit{GMM expansion.} To assign outlier frames to nearby clusters, we modeled each HDBSCAN cluster as a multivariate Gaussian distribution. For each cluster $k$, we estimated the mean $\boldsymbol{\mu}_k$ and covariance $\boldsymbol{\Sigma}_k$ from the assigned frames, and computed a reference probability density $p_k^{\mathrm{ref}}$ as the 95th percentile of the probability density evaluated at cluster members. Each outlier frame $\mathbf{x}$ was then assigned to the cluster $k^*$ maximizing the relative probability up to an expansion threshold ($\varepsilon = 0.01)$. Outliers not meeting this criterion remained unassigned.

\textit{Cluster curation.} To focus the analysis on structurally interpretable states, we excluded clusters corresponding to heterogeneous unfolded ensembles. Specifically, small unfolded clusters were excluded for 1FME (2 clusters) and 2F4K (3 clusters), while large unfolded clusters were excluded for A3D (1 cluster).

\paragraph{Cross-Correlation Matrices}
To characterize correlated motions within each conformational state, we computed two complementary correlation matrices: the Dynamic Cross-Correlation Map (DCCM) based on atomic displacements, and the Dynamic Cross-Orientation Map (DCOM) based on local coordinate system orientations.

\textit{Dynamic Cross-Correlation Map (DCCM).} The DCCM quantifies correlated translational motions between residue pairs \cite{Ichiye1991}. For a trajectory of $\mathrm{C\alpha}$ coordinates, let $\Delta \mathbf{r}_i(t) = \mathbf{r}_i(t) - \langle \mathbf{r}_i \rangle$ denote the displacement of residue $i$ from its mean position. The DCCM element $C_{ij}$ is defined as the normalized covariance of displacements:
\begin{equation}
    C_{ij}^{\mathrm{DCCM}} = \frac{\langle \Delta \mathbf{r}_i \cdot \Delta \mathbf{r}_j \rangle}{\sqrt{\langle |\Delta \mathbf{r}_i|^2 \rangle \langle |\Delta \mathbf{r}_j|^2 \rangle}},
\end{equation}
where $\langle \cdot \rangle$ denotes the time average over frames within a given cluster. Values range from $-1$ (perfectly anti-correlated motion) to $+1$ (perfectly correlated motion), with $C_{ij}^{\mathrm{DCCM}} = 0$ indicating uncorrelated motion.

\textit{Dynamic Cross-Orientation Map (DCOM).} The DCOM captures correlated rotational motions by comparing the orientations of local coordinate systems. For each residue $i$, let $\mathbf{n}_i(t)$ denote the normal vector of the local coordinate system (the second column of $\mathrm{LCS}_{(i,t)}$, orthogonal to the peptide plane). The DCOM element $C_{ij}$ is defined as the average relative orientation between residue pairs:
\begin{equation}
    C_{ij}^{\mathrm{DCOM}} = \mathrm{atan2}\left( \langle (\mathbf{n}_i \times \mathbf{n}_j) \cdot \mathbf{e} \rangle, \langle \mathbf{n}_i \cdot \mathbf{n}_j \rangle \right),
\end{equation}
where $\mathbf{e}$ is a fixed reference axis (here, $\mathbf{e} = [1, 0, 0]^\top$) and $\mathrm{atan2}$ returns the signed angle in degrees. The DCOM thus reports the mean dihedral-like angle between the peptide planes of residue pairs, ranging from $-180^\circ$ to $+180^\circ$.
The reference axis $\mathbf{e}$ establishes a consistent coordinate frame for computing signed dihedral-like angles between peptide plane normals.
While the absolute DCOM values and their differences depend on the choice of $\mathbf{e}$, the qualitative interpretation--whether relative orientations between residue pairs differ between clusters--is axis-independent. Applying a uniform reference axis throughout the analysis ensures internal consistency. Cluster differences are computed as $\Delta C_{ij} = ((C_{ij}^B - C_{ij}^A) + 180) \mod 360)-180$, ensuring proper handling of angular periodicity.

\subsection*{SARS-CoV-2 nsp13}

For the SARS-CoV-2 nsp13 helicase analysis, we assessed whether the geometric similarities encoded by each feature representation reflect physically meaningful structural similarities. The analysis comprises four components: (1) validation of feature space geometry against structural metrics, (2) Gram matrix analysis to characterize the similarity structure, (3) dimensionality reduction via PCA, and (4) clustering evaluation using multiple concordance metrics.

\paragraph{Pairwise RMSD and lDDT}
To establish ground-truth structural similarities, we computed pairwise Root Mean Square Deviation (RMSD) and local Distance Difference Test (lDDT) scores between all trajectory frames.

\textit{RMSD.} Pairwise RMSD was computed on $\mathrm{C\alpha}$ atoms after optimal superposition. RMSD provides a global measure of structural dissimilarity, sensitive to both local and large-scale conformational changes.

\textit{lDDT.} The local Distance Difference Test~\cite{Adhikari2021} evaluates structural similarity based on the preservation of local inter-residue distances, without requiring superposition. For each pair of residues $(i, j)$ within a distance threshold $R_0$ in the reference structure, lDDT assesses whether the corresponding distance in the target structure is preserved within tolerance thresholds of 0.5, 1.0, 2.0, and 4.0~\AA:
\begin{equation}
    \mathrm{lDDT} = \frac{1}{4|S|} \sum_{(i,j) \in S} \sum_{\delta \in \{0.5, 1, 2, 4\}} \mathbf{1}\left[ |d_{ij} - d'_{ij}| < \delta \right],
\end{equation}
where $S = \{(i,j) : d_{ij} < R_0\}$ is the set of residue pairs in contact (here, $R_0 = 15 ~\mathring{A}$), $d_{ij}$ and $d'_{ij}$ are the inter-residue distances in the reference and target structures, and $\mathbf{1}[\cdot]$ is the indicator function. lDDT ranges from 0 to 1, with higher values indicating better preservation of local geometry. Unlike RMSD, lDDT is insensitive to global rigid-body motions and emphasizes local structural accuracy.

\paragraph{Gram Matrices Analysis}
A central question in designing feature representations is whether similarities in feature space correspond to physically meaningful structural similarities. To address this, we analyzed the Gram matrices induced by each representation and compared them against the structural similarity metrics.

\textit{Gram Matrices.} For a feature matrix $F \in \mathbb{R}^{T \times d}$ containing $T$ frames with $d$-dimensional features, the Gram matrix $G \in \mathbb{R}^{T \times T}$ is defined as:
\begin{equation}
    G_{ij} = \langle \mathbf{f}_i, \mathbf{f}_j \rangle = \sum_{k=1}^{d} f_{ik} f_{jk},
\end{equation}
where $\mathbf{f}_i$ denotes the feature vector for frame $i$. The Gram matrix encodes all pairwise inner products in feature space; frames with similar features yield large positive entries, while dissimilar frames yield small or negative entries. Importantly, the Gram matrix fully determines the geometry of the data up to rotation, i.e., any distance or angle in feature space can be recovered from $G$.

\textit{Rank-1 Decomposition.} To isolate the dominant modes of variation captured by each representation, we performed eigendecomposition of the Gram matrices. For a Gram matrix $G = Q \Lambda Q^\top$ with eigenvalues $\lambda_1 \geq \lambda_2 \geq \cdots$ and eigenvectors $\mathbf{q}_1, \mathbf{q}_2, \ldots$, the rank-1 components are:
\begin{equation}
    G^{(k)} = \lambda_k \mathbf{q}_k \mathbf{q}_k^\top.
\end{equation}
Each rank-1 component captures the contribution of a single mode of variation to the overall similarity structure. By examining the correlation of individual components with structural metrics, we can assess whether the dominant modes correspond to physically meaningful conformational changes or to spurious variation.

\textit{Correlation.} To quantify how well the feature space geometry reflects structural similarity, we computed Spearman rank correlations between the lower-triangular elements of each Gram matrix and the corresponding pairwise RMSD and lDDT values. A strong negative correlation with RMSD (or positive correlation with lDDT) indicates that the feature representation faithfully encodes structural similarity, i.e., frames that are close in feature space are also structurally similar.

\paragraph{Principal Component Analysis}
PCA was performed on each feature representation to obtain a low-dimensional embedding for visualization and clustering. For each representation, we retained the first two principal components, capturing the dominant modes of conformational variation.

\paragraph{Clustering}
Conformational states were identified by clustering in the two-dimensional PCA embedding, and the quality of the resulting partitions was evaluated using metrics that assess both internal cohesion and cross-representation consistency.

\textit{HDBSCAN.} Clustering was performed using HDBSCAN~\cite{Campello2013} with a minimum cluster size of 100 frames and a cluster selection epsilon of 0.01. Cluster labels were sorted in descending order by cluster size, and cluster centroids were computed as the mean position of all frames assigned to each cluster.

\textit{Silhouette Score.} The silhouette score quantifies how well each frame fits within its assigned cluster relative to other clusters, using the pairwise RMSD matrix as the distance metric. The silhouette score ranges from $-1$ to $+1$: values near $+1$ indicate that the frame is well-matched to its cluster and poorly matched to neighboring clusters, values near $0$ indicate the frame lies on the boundary between clusters, and negative values suggest potential misassignment \cite{Rousseeuw1987,scikit-learn}. The overall silhouette score is the mean over all frames. By computing the silhouette score using RMSD as the distance metric rather than Euclidean distance in feature space, we directly assess whether the clustering partitions the trajectory into structurally coherent states.

\textit{Adjusted Mutual Information.} To assess the consistency of cluster assignments across different feature representations, we computed the Adjusted Mutual Information (AMI) between all pairs of clusterings. AMI equals 1 for identical clusterings, 0 for independent clusterings (no better than chance), and can be negative if the agreement is worse than expected by chance. Only frames assigned to clusters in both representations (i.e., excluding outliers) were included in the comparison \cite{Vinh2009,scikit-learn}.

\textit{Adjusted Rand Index.} The Adjusted Rand Index (ARI) provides an alternative measure of clustering agreement based on pairwise co-assignments. For two clusterings, let $a$ be the number of frame pairs that are in the same cluster in both clusterings, and $b$ the number of pairs in different clusters in both. Like AMI, ARI equals 1 for identical clusterings and 0 for chance-level agreement \cite{Hubert1985,scikit-learn}. ARI emphasizes agreement on pairwise relationships, while AMI emphasizes agreement on cluster membership distributions.

\subsection*{Protein-Protein Association}

For the protein-protein association analysis, we evaluated whether internal conformational features of individual monomers contain information predictive of the binding state. The analysis comprises three components: (1) definition of the bound/unbound state based on interface RMSD, (2) hierarchical dimensionality reduction via two-step PCA, and (3) quantification of state-predictive information via MI.

\paragraph{IRMSD}
To define the bound and unbound states, we computed the Interface Root Mean Square Deviation (IRMSD) between each trajectory frame and the crystallographic bound-state structure.

Interface residues were identified from the solved crystal structure as those $\mathrm{C\alpha}$ atoms within 10~\AA{} of any $\mathrm{C\alpha}$ atom on the partner chain. For each trajectory frame, the interface atoms from both monomers were extracted, superimposed onto the corresponding atoms in the crystal structure, and the IRMSD was computed. Unlike global RMSD, IRMSD specifically measures the accuracy of the interface geometry, making it well-suited for distinguishing bound from unbound or encounter complexes.

For the Insulin dimer (1) and RBD-ACE2 systems, IRMSD was not suitable for state classification. The Insulin dimer (1) trajectory contained a non-canonical bound state with an interface geometry differing substantially from the crystal structure, while the RBD-ACE2 system required a more general association metric. For these systems, we used the distance between the centers of geometry ($d_\mathrm{COG}$) of the two monomers. This metric distinguishes associated from dissociated states regardless of the specific interface geometry.

\textit{Interaction Threshold.} To classify frames as bound or unbound, we determined a threshold from the empirical distribution of the association metric (IRMSD or $d_{\mathrm{COG}}$). We computed the Probability Density Function (PDF) using kernel density estimation \cite{scipy} and identified the threshold as the value at which the gradient of the PDF reaches its minimum; this corresponds to the boundary between the bound-state and unbound-state populations. Frames below this threshold were labeled as bound (label = 1), and frames above as unbound (label = 0).

\paragraph{Two-Step Principal Component Analysis}
To identify collective variables that capture binding-relevant conformational changes, we applied a hierarchical PCA procedure that first reduces each monomer independently, then combines the reduced representations.

\textit{Step 1: Per-monomer PCA.} For each monomer, the feature matrix was subjected to PCA with whitening. The first two principal components were retained, capturing the dominant modes of internal conformational variation for each monomer independently. Whitening ensures that the retained components have unit variance and are uncorrelated, providing a normalized basis for subsequent analysis.

\textit{Step 2: Combined PCA.} The per-monomer principal components were concatenated to form a 4-dimensional representation (2 components $\times$ 2 monomers). A second PCA with whitening was then applied to this combined representation, and the first two principal components were retained. This hierarchical approach identifies collective modes of variation that span both monomers, potentially capturing coordinated conformational changes associated with binding.

\textit{Contribution.} To identify which residues contribute most to each principal component, we computed residue-level contributions by propagating the loadings through both PCA stages. For a residue $r$ with original features indexed by $k \in \mathcal{K}_r$, the contribution to the $i$-th combined principal component is:
\begin{equation}
    C_r^{(i)} = \frac{\displaystyle\sum_{k \in \mathcal{K}_r} \left( \sum_{j=1}^{2} w_{kj}^{(1)} w_{ji}^{(2)} \right)^2}{\displaystyle\sum_{r'} \sum_{k \in \mathcal{K}_{r'}} \left( \sum_{j=1}^{2} w_{kj}^{(1)} w_{ji}^{(2)} \right)^2},
\end{equation}
where $w_{kj}^{(1)}$ are the loadings from the first-stage PCA mapping original features to per-monomer components, and $w_{ji}^{(2)}$ are the loadings from the second-stage PCA mapping concatenated per-monomer components to combined components. This formulation identifies residues whose conformational fluctuations are most strongly coupled to the dominant collective modes.

\textit{Mutual Information.} To quantify how well each representation separates bound and unbound states, we computed the MI between the first principal component and the binary binding label. MI measures the reduction in uncertainty about the binding state given knowledge of the principal component value. Higher MI indicates that the representation captures conformational features that are more strongly associated with the binding state. MI was estimated using $k$-nearest neighbor methods as implemented in scikit-learn~\cite{scikit-learn}, with continuous features and discrete labels.

\paragraph{Clustering (RBD-ACE2)}
For the RBD-ACE2 system, we additionally performed hierarchical clustering to identify distinct conformational states within the trajectory.

\textit{Ward Clustering.} Agglomerative hierarchical clustering was performed on the two-dimensional PCA embedding using Ward's minimum variance method~\cite{Ward1963,Husic2017}. The hierarchical tree was cut at a distance threshold of 10 to obtain flat cluster assignments. Final cluster labels were sorted in descending order by cluster size.

\subsection*{ManiProt Library}
To characterize the computational scaling of the proposed methods, we profiled the core operations of the ManiProt library across a range of problem sizes.

\paragraph{Benchmarking Protocol}
Synthetic data were generated for systematic profiling. For the $\mathrm{SO}(3)$ logarithmic map, random rotation matrices were constructed from uniformly sampled angles. For the Pointcloud logarithmic map, random three-dimensional point clouds were generated from a standard normal distribution. The number of samples (trajectory frames) was varied as powers of two from $2^0$ to $2^{19}$ (1 to 524,288), and the number of residues was varied from $2^2$ to $2^9$ (4 to 512). Each configuration was evaluated over three independent replicas, and the mean execution time was recorded.

\paragraph{Profiled Operations}
Four operations were profiled:
\begin{enumerate}
    \item \textbf{SO(3) logarithmic map}: Projection of rotation matrices onto the Lie algebra, computed per-residue across all frames.
    \item \textbf{Pointcloud logarithmic map}: Projection of point clouds onto the tangent space at a reference configuration.
    \item \textbf{Metric tensor computation}: Construction of the Riemannian metric tensor for Pointcloud features.
    \item \textbf{Metric tensor inversion}: Eigendecomposition and pseudo-inversion of the metric tensor, required for computing the natural gradient.
\end{enumerate}

Execution times were normalized relative to the smallest problem size and visualized on a logarithmic scale. All computations were performed using Numba-accelerated implementations with parallel execution enabled.

\backmatter

\bmhead{Supplementary information}

Figures S1 to S23\\
Table S1

\bmhead{Acknowledgements}

\paragraph*{Funding:}
This work was supported by funds from the Swiss National Science Foundation (PCEFP3 194606)
\paragraph*{Author contributions:}
Conceptualization: AG, TL. 
Methodology: AG.
Investigation: AG.
Visualization: AG.
Supervision: TL.
Writing --- original draft: AG.
Writing --- review \& editing: AG, TL.
\paragraph*{Competing interests:}
There are no competing interests to declare.
\paragraph*{Data and materials availability:}
The ManiProt library is available at \textit{https://github.com/ibmm-unibe-ch/ManiProt}. Computed features used in this study are publicly accessible via Zenodo at \textit{https://doi.org/10.5281/zenodo.18712231}.

\clearpage





\bibliography{bibliography}

@article{Ward1963,
  title = {Hierarchical Grouping to Optimize an Objective Function},
  volume = {58},
  ISSN = {1537-274X},
  url = {http://dx.doi.org/10.1080/01621459.1963.10500845},
  DOI = {10.1080/01621459.1963.10500845},
  number = {301},
  journal = {Journal of the American Statistical Association},
  publisher = {Informa UK Limited},
  author = {Ward,  Joe H.},
  year = {1963},
  month = mar,
  pages = {236–244}
}

@article{Kabsch1983,
  title = {Dictionary of protein secondary structure: Pattern recognition of hydrogen‐bonded and geometrical features},
  volume = {22},
  ISSN = {1097-0282},
  url = {http://dx.doi.org/10.1002/bip.360221211},
  DOI = {10.1002/bip.360221211},
  number = {12},
  journal = {Biopolymers},
  publisher = {Wiley},
  author = {Kabsch,  Wolfgang and Sander,  Christian},
  year = {1983},
  month = dec,
  pages = {2577–2637}
}

@article{Hubert1985,
  title = {Comparing partitions},
  volume = {2},
  ISSN = {1432-1343},
  url = {http://dx.doi.org/10.1007/BF01908075},
  DOI = {10.1007/bf01908075},
  number = {1},
  journal = {Journal of Classification},
  publisher = {Springer Science and Business Media LLC},
  author = {Hubert,  Lawrence and Arabie,  Phipps},
  year = {1985},
  month = dec,
  pages = {193–218}
}

@article{Rousseeuw1987,
  title = {Silhouettes: A graphical aid to the interpretation and validation of cluster analysis},
  volume = {20},
  ISSN = {0377-0427},
  url = {http://dx.doi.org/10.1016/0377-0427(87)90125-7},
  DOI = {10.1016/0377-0427(87)90125-7},
  journal = {Journal of Computational and Applied Mathematics},
  publisher = {Elsevier BV},
  author = {Rousseeuw,  Peter J.},
  year = {1987},
  month = nov,
  pages = {53–65}
}

@article{Ichiye1991,
  title = {Collective motions in proteins: A covariance analysis of atomic fluctuations in molecular dynamics and normal mode simulations},
  volume = {11},
  ISSN = {1097-0134},
  url = {http://dx.doi.org/10.1002/prot.340110305},
  DOI = {10.1002/prot.340110305},
  number = {3},
  journal = {Proteins: Structure,  Function,  and Bioinformatics},
  publisher = {Wiley},
  author = {Ichiye,  Toshiko and Karplus,  Martin},
  year = {1991},
  month = nov,
  pages = {205–217}
}

@article{Johansson1997,
  title = {Solution structure of the albumin-binding GA module: a versatile bacterial protein domain},
  volume = {266},
  ISSN = {0022-2836},
  url = {http://dx.doi.org/10.1006/jmbi.1996.0856},
  DOI = {10.1006/jmbi.1996.0856},
  number = {5},
  journal = {Journal of Molecular Biology},
  publisher = {Elsevier BV},
  author = {Johansson,  Maria U and de Ch\^ateau,  Maarten and Wikstr\"{o}m,  Mats and Forsén,  Sture and Drakenberg,  Torbj\"{o}rn and Bj\"{o}rck,  Lars},
  year = {1997},
  month = mar,
  pages = {859–865}
}

@article{Walsh1999,
  title = {Solution structure and dynamics of a
            de novo
            designed three-helix bundle protein},
  volume = {96},
  ISSN = {1091-6490},
  url = {http://dx.doi.org/10.1073/pnas.96.10.5486},
  DOI = {10.1073/pnas.96.10.5486},
  number = {10},
  journal = {Proceedings of the National Academy of Sciences},
  publisher = {Proceedings of the National Academy of Sciences},
  author = {Walsh,  Scott T. R. and Cheng,  Hong and Bryson,  James W. and Roder,  Heinrich and DeGrado,  William F.},
  year = {1999},
  month = may,
  pages = {5486–5491}
}

@book{Hyvrinen2001,
  title = {Independent Component Analysis},
  ISBN = {9780471221319},
  url = {http://dx.doi.org/10.1002/0471221317},
  DOI = {10.1002/0471221317},
  publisher = {Wiley},
  author = {Hyv\"{a}rinen,  Aapo and Karhunen,  Juha and Oja,  Erkki},
  year = {2001},
  month = may 
}

@book{Berger2003,
  title = {A Panoramic View of Riemannian Geometry},
  ISBN = {9783642182457},
  url = {http://dx.doi.org/10.1007/978-3-642-18245-7},
  DOI = {10.1007/978-3-642-18245-7},
  publisher = {Springer Berlin Heidelberg},
  author = {Berger,  Marcel},
  year = {2003}
}

@article{Sarisky2001,
  title = {The $\beta\beta\alpha$ fold: explorations in sequence space11Edited by M. F. Summers},
  volume = {307},
  ISSN = {0022-2836},
  url = {http://dx.doi.org/10.1006/jmbi.2000.4345},
  DOI = {10.1006/jmbi.2000.4345},
  number = {5},
  journal = {Journal of Molecular Biology},
  publisher = {Elsevier BV},
  author = {Sarisky,  Catherine A and Mayo,  Stephen L},
  year = {2001},
  month = apr,
  pages = {1411–1418}
}

@article{Fletcher2004,
  title={Principal geodesic analysis for the study of nonlinear statistics of shape},
  author={Fletcher, P Thomas and Lu, Conglin and Pizer, Stephen M and Joshi, Sarang},
  journal={IEEE transactions on medical imaging},
  volume={23},
  number={8},
  pages={995--1005},
  year={2004},
  publisher={IEEE}
}

@inproceedings{Vinh2009,
  series = {ICML ’09},
  title = {Information theoretic measures for clusterings comparison: is a correction for chance necessary?},
  url = {http://dx.doi.org/10.1145/1553374.1553511},
  DOI = {10.1145/1553374.1553511},
  booktitle = {Proceedings of the 26th Annual International Conference on Machine Learning},
  publisher = {ACM},
  author = {Vinh,  Nguyen Xuan and Epps,  Julien and Bailey,  James},
  year = {2009},
  month = jun,
  collection = {ICML ’09}
}

@article{Pande2010,
  title={Folding@ home},
  author={Pande, Vijay and others},
  journal={Distributed computing},
  year={2010}
}

@article{LindorffLarsen2011,
  title = {How Fast-Folding Proteins Fold},
  volume = {334},
  ISSN = {1095-9203},
  url = {http://dx.doi.org/10.1126/science.1208351},
  DOI = {10.1126/science.1208351},
  number = {6055},
  journal = {Science},
  publisher = {American Association for the Advancement of Science (AAAS)},
  author = {Lindorff-Larsen,  Kresten and Piana,  Stefano and Dror,  Ron O. and Shaw,  David E.},
  year = {2011},
  month = oct,
  pages = {517–520}
}

@inbook{Campello2013,
  title = {Density-Based Clustering Based on Hierarchical Density Estimates},
  ISBN = {9783642374562},
  ISSN = {1611-3349},
  url = {http://dx.doi.org/10.1007/978-3-642-37456-2_14},
  DOI = {10.1007/978-3-642-37456-2_14},
  booktitle = {Advances in Knowledge Discovery and Data Mining},
  publisher = {Springer Berlin Heidelberg},
  author = {Campello,  Ricardo J. G. B. and Moulavi,  Davoud and Sander,  Joerg},
  year = {2013},
  pages = {160–172}
}

@article{PrezHernndez2013,
  title = {Identification of slow molecular order parameters for Markov model construction},
  volume = {139},
  ISSN = {1089-7690},
  url = {http://dx.doi.org/10.1063/1.4811489},
  DOI = {10.1063/1.4811489},
  number = {1},
  journal = {The Journal of Chemical Physics},
  publisher = {AIP Publishing},
  author = {Pérez-Hernández,  Guillermo and Paul,  Fabian and Giorgino,  Toni and De Fabritiis,  Gianni and Noé,  Frank},
  year = {2013},
  month = jul 
}

@inproceedings{Shaw2014,
  title = {Anton 2: Raising the Bar for Performance and Programmability in a Special-Purpose Molecular Dynamics Supercomputer},
  url = {http://dx.doi.org/10.1109/sc.2014.9},
  DOI = {10.1109/sc.2014.9},
  booktitle = {SC14: International Conference for High Performance Computing,  Networking,  Storage and Analysis},
  publisher = {IEEE},
  author = {Shaw,  David E. and Grossman,  J.P. and Bank,  Joseph A. and Batson,  Brannon and Butts,  J. Adam and Chao,  Jack C. and Deneroff,  Martin M. and Dror,  Ron O. and Even,  Amos and Fenton,  Christopher H. and Forte,  Anthony and Gagliardo,  Joseph and Gill,  Gennette and Greskamp,  Brian and Ho,  C. Richard and Ierardi,  Douglas J. and Iserovich,  Lev and Kuskin,  Jeffrey S. and Larson,  Richard H. and Layman,  Timothy and Lee,  Li-Siang and Lerer,  Adam K. and Li,  Chester and Killebrew,  Daniel and Mackenzie,  Kenneth M. and Mok,  Shark Yeuk-Hai and Moraes,  Mark A. and Mueller,  Rolf and Nociolo,  Lawrence J. and Peticolas,  Jon L. and Quan,  Terry and Ramot,  Daniel and Salmon,  John K. and Scarpazza,  Daniele P. and Schafer,  U. Ben and Siddique,  Naseer and Snyder,  Christopher W. and Spengler,  Jochen and Tang,  Ping Tak Peter and Theobald,  Michael and Toma,  Horia and Towles,  Brian and Vitale,  Benjamin and Wang,  Stanley C. and Young,  Cliff},
  year = {2014},
  month = nov,
  pages = {41–53}
}

@article{Husic2017,
  title = {Ward Clustering Improves Cross-Validated Markov State Models of Protein Folding},
  volume = {13},
  ISSN = {1549-9626},
  url = {http://dx.doi.org/10.1021/acs.jctc.6b01238},
  DOI = {10.1021/acs.jctc.6b01238},
  number = {3},
  journal = {Journal of Chemical Theory and Computation},
  publisher = {American Chemical Society (ACS)},
  author = {Husic,  Brooke E. and Pande,  Vijay S.},
  year = {2017},
  month = feb,
  pages = {963–967}
}

@article{Ingraham2019,
  title={Generative models for graph-based protein design},
  author={Ingraham, John and Garg, Vikas and Barzilay, Regina and Jaakkola, Tommi},
  journal={Advances in neural information processing systems},
  volume={32},
  year={2019}
}

@article{Pan2019,
  title = {Atomic-level characterization of protein–protein association},
  volume = {116},
  ISSN = {1091-6490},
  url = {http://dx.doi.org/10.1073/pnas.1815431116},
  DOI = {10.1073/pnas.1815431116},
  number = {10},
  journal = {Proceedings of the National Academy of Sciences},
  publisher = {Proceedings of the National Academy of Sciences},
  author = {Pan,  Albert C. and Jacobson,  Daniel and Yatsenko,  Konstantin and Sritharan,  Duluxan and Weinreich,  Thomas M. and Shaw,  David E.},
  year = {2019},
  month = feb,
  pages = {4244–4249}
}

@article{Wu2019,
  title = {Variational Approach for Learning Markov Processes from Time Series Data},
  volume = {30},
  ISSN = {1432-1467},
  url = {http://dx.doi.org/10.1007/s00332-019-09567-y},
  DOI = {10.1007/s00332-019-09567-y},
  number = {1},
  journal = {Journal of Nonlinear Science},
  publisher = {Springer Science and Business Media LLC},
  author = {Wu,  Hao and Noé,  Frank},
  year = {2019},
  month = aug,
  pages = {23–66}
}

@misc{Blanco2021,
  doi = {10.48550/ARXIV.2103.15980},
  url = {https://arxiv.org/abs/2103.15980},
  author = {Blanco-Claraco,  José Luis},
  title = {A tutorial on $\mathbf{SE}(3)$ transformation parameterizations and on-manifold optimization},
  publisher = {arXiv},
  year = {2021},
  copyright = {Creative Commons Attribution Non Commercial Share Alike 4.0 International}
}

@article{Adhikari2021,
  title = {DISTEVAL: a web server for evaluating predicted protein distances},
  volume = {22},
  ISSN = {1471-2105},
  url = {http://dx.doi.org/10.1186/s12859-020-03938-z},
  DOI = {10.1186/s12859-020-03938-z},
  number = {1},
  journal = {BMC Bioinformatics},
  publisher = {Springer Science and Business Media LLC},
  author = {Adhikari,  Badri and Shrestha,  Bikash and Bernardini,  Matthew and Hou,  Jie and Lea,  Jamie},
  year = {2021},
  month = jan 
}

@article{Glielmo2021,
  title = {Unsupervised Learning Methods for Molecular Simulation Data},
  volume = {121},
  ISSN = {1520-6890},
  url = {http://dx.doi.org/10.1021/acs.chemrev.0c01195},
  DOI = {10.1021/acs.chemrev.0c01195},
  number = {16},
  journal = {Chemical Reviews},
  publisher = {American Chemical Society (ACS)},
  author = {Glielmo,  Aldo and Husic,  Brooke E. and Rodriguez,  Alex and Clementi,  Cecilia and Noé,  Frank and Laio,  Alessandro},
  year = {2021},
  month = may,
  pages = {9722–9758}
}

@article{Jumper2021,
  title={Highly accurate protein structure prediction with AlphaFold},
  author={Jumper, John and Evans, Richard and Pritzel, Alexander and Green, Tim and Figurnov, Michael and Ronneberger, Olaf and Tunyasuvunakool, Kathryn and Bates, Russ and {\v{Z}}{\'\i}dek, Augustin and Potapenko, Anna and others},
  journal={nature},
  volume={596},
  number={7873},
  pages={583--589},
  year={2021},
  publisher={Nature Publishing Group}
}

@article{Morris2021,
  title = {Using molecular docking and molecular dynamics to investigate protein-ligand interactions},
  volume = {35},
  ISSN = {1793-6640},
  url = {http://dx.doi.org/10.1142/s0217984921300027},
  DOI = {10.1142/s0217984921300027},
  number = {08},
  journal = {Modern Physics Letters B},
  publisher = {World Scientific Pub Co Pte Ltd},
  author = {Morris,  Connor J. and Corte,  Dennis Della},
  year = {2021},
  month = feb,
  pages = {2130002}
}

@misc{Nurlanov2021,
  title = {Exploring SO(3) Logarithmic Map: Degeneracies and Derivatives},
  author = {Nurlanov, Zhakshylyk},
  year = {2021}
}

@InProceedings{Satorras2021,
  title = 	 {E(n) Equivariant Graph Neural Networks},
  author =       {Satorras, V\'{\i}ctor Garcia and Hoogeboom, Emiel and Welling, Max},
  booktitle = 	 {Proceedings of the 38th International Conference on Machine Learning},
  pages = 	 {9323--9332},
  year = 	 {2021},
  editor = 	 {Meila, Marina and Zhang, Tong},
  volume = 	 {139},
  series = 	 {Proceedings of Machine Learning Research},
  month = 	 {18--24 Jul},
  publisher =    {PMLR},
  url = 	 {https://proceedings.mlr.press/v139/satorras21a.html}
}

@inproceedings{Shaw2021,
  series = {SC ’21},
  title = {Anton 3: twenty microseconds of molecular dynamics simulation before lunch},
  url = {http://dx.doi.org/10.1145/3458817.3487397},
  DOI = {10.1145/3458817.3487397},
  booktitle = {Proceedings of the International Conference for High Performance Computing,  Networking,  Storage and Analysis},
  publisher = {ACM},
  author = {Shaw,  David E. and Adams,  Peter J. and Azaria,  Asaph and Bank,  Joseph A. and Batson,  Brannon and Bell,  Alistair and Bergdorf,  Michael and Bhatt,  Jhanvi and Butts,  J. Adam and Correia,  Timothy and Dirks,  Robert M. and Dror,  Ron O. and Eastwood,  Michael P. and Edwards,  Bruce and Even,  Amos and Feldmann,  Peter and Fenn,  Michael and Fenton,  Christopher H. and Forte,  Anthony and Gagliardo,  Joseph and Gill,  Gennette and Gorlatova,  Maria and Greskamp,  Brian and Grossman,  J.P. and Gullingsrud,  Justin and Harper,  Anissa and Hasenplaugh,  William and Heily,  Mark and Heshmat,  Benjamin Colin and Hunt,  Jeremy and Ierardi,  Douglas J. and Iserovich,  Lev and Jackson,  Bryan L. and Johnson,  Nick P. and Kirk,  Mollie M. and Klepeis,  John L. and Kuskin,  Jeffrey S. and Mackenzie,  Kenneth M. and Mader,  Roy J. and McGowen,  Richard and McLaughlin,  Adam and Moraes,  Mark A. and Nasr,  Mohamed H. and Nociolo,  Lawrence J. and O’Donnell,  Lief and Parker,  Andrew and Peticolas,  Jon L. and Pocina,  Goran and Predescu,  Cristian and Quan,  Terry and Salmon,  John K. and Schwink,  Carl and Shim,  Keun Sup and Siddique,  Naseer and Spengler,  Jochen and Szalay,  Tamas and Tabladillo,  Raymond and Tartler,  Reinhard and Taube,  Andrew G. and Theobald,  Michael and Towles,  Brian and Vick,  William and Wang,  Stanley C. and Wazlowski,  Michael and Weingarten,  Madeleine J. and Williams,  John M. and Yuh,  Kevin A.},
  year = {2021},
  month = nov,
  pages = {1–11},
  collection = {SC ’21}
}

@article{Zimmerman2021,
  title = {SARS-CoV-2 Simulations go Exascale to Capture Spike Opening and Reveal Cryptic Pockets Across the Proteome},
  volume = {120},
  ISSN = {0006-3495},
  url = {http://dx.doi.org/10.1016/j.bpj.2020.11.1909},
  DOI = {10.1016/j.bpj.2020.11.1909},
  number = {3},
  journal = {Biophysical Journal},
  publisher = {Elsevier BV},
  author = {Zimmerman,  Maxwell I. and Bowman,  Gregory},
  year = {2021},
  month = feb,
  pages = {299a}
}

@article{Chen2022,
  title = {Ensemble cryo-EM reveals conformational states of the nsp13 helicase in the SARS-CoV-2 helicase replication–transcription complex},
  volume = {29},
  ISSN = {1545-9985},
  url = {http://dx.doi.org/10.1038/s41594-022-00734-6},
  DOI = {10.1038/s41594-022-00734-6},
  number = {3},
  journal = {Nature Structural \&; Molecular Biology},
  publisher = {Springer Science and Business Media LLC},
  author = {Chen,  James and Wang,  Qi and Malone,  Brandon and Llewellyn,  Eliza and Pechersky,  Yakov and Maruthi,  Kashyap and Eng,  Ed T. and Perry,  Jason K. and Campbell,  Elizabeth A. and Shaw,  David E. and Darst,  Seth A.},
  year = {2022},
  month = mar,
  pages = {250–260}
}

@misc{Diepeveen2023,
  doi = {10.48550/ARXIV.2308.07818},
  url = {https://arxiv.org/abs/2308.07818},
  author = {Diepeveen,  Willem and Esteve-Yag\"{u}e,  Carlos and Lellmann,  Jan and \"{O}ktem,  Ozan and Sch\"{o}nlieb,  Carola-Bibiane},
  title = {Riemannian geometry for efficient analysis of protein dynamics data},
  publisher = {arXiv},
  year = {2023},
  copyright = {Creative Commons Attribution Non Commercial No Derivatives 4.0 International}
}

@inproceedings{Klein2023,
 author = {Klein, Leon and Kr\"{a}mer, Andreas and Noe, Frank},
 booktitle = {Advances in Neural Information Processing Systems},
 editor = {A. Oh and T. Naumann and A. Globerson and K. Saenko and M. Hardt and S. Levine},
 pages = {59886--59910},
 publisher = {Curran Associates, Inc.},
 title = {Equivariant flow matching},
 url = {https://proceedings.neurips.cc/paper_files/paper/2023/file/bc827452450356f9f558f4e4568d553b-Paper-Conference.pdf},
 volume = {36},
 year = {2023}
}

@article{Voelz2023,
  title = {Folding@home: Achievements from over 20 years of citizen science herald the exascale era},
  volume = {122},
  ISSN = {0006-3495},
  url = {http://dx.doi.org/10.1016/j.bpj.2023.03.028},
  DOI = {10.1016/j.bpj.2023.03.028},
  number = {14},
  journal = {Biophysical Journal},
  publisher = {Elsevier BV},
  author = {Voelz,  Vincent A. and Pande,  Vijay S. and Bowman,  Gregory R.},
  year = {2023},
  month = jul,
  pages = {2852–2863}
}

@article{numpy,
  title = {Array programming with NumPy},
  volume = {585},
  ISSN = {1476-4687},
  url = {http://dx.doi.org/10.1038/s41586-020-2649-2},
  DOI = {10.1038/s41586-020-2649-2},
  number = {7825},
  journal = {Nature},
  publisher = {Springer Science and Business Media LLC},
  author = {Harris,  Charles R. and Millman,  K. Jarrod and van der Walt,  Stéfan J. and Gommers,  Ralf and Virtanen,  Pauli and Cournapeau,  David and Wieser,  Eric and Taylor,  Julian and Berg,  Sebastian and Smith,  Nathaniel J. and Kern,  Robert and Picus,  Matti and Hoyer,  Stephan and van Kerkwijk,  Marten H. and Brett,  Matthew and Haldane,  Allan and del Río,  Jaime Fernández and Wiebe,  Mark and Peterson,  Pearu and Gérard-Marchant,  Pierre and Sheppard,  Kevin and Reddy,  Tyler and Weckesser,  Warren and Abbasi,  Hameer and Gohlke,  Christoph and Oliphant,  Travis E.},
  year = {2020},
  month = sep,
  pages = {357–362}
}

@inproceedings{numba,
  series = {SC15},
  title = {Numba: a LLVM-based Python JIT compiler},
  url = {http://dx.doi.org/10.1145/2833157.2833162},
  DOI = {10.1145/2833157.2833162},
  booktitle = {Proceedings of the Second Workshop on the LLVM Compiler Infrastructure in HPC},
  publisher = {ACM},
  author = {Lam,  Siu Kwan and Pitrou,  Antoine and Seibert,  Stanley},
  year = {2015},
  month = nov,
  pages = {1–6},
  collection = {SC15}
}

@article{deeptime,
  title={Deeptime: a Python library for machine learning dynamical models from time series data},
  author={Hoffmann, Moritz and Scherer, Martin Konrad and Hempel, Tim and Mardt, Andreas and de Silva, Brian and Husic, Brooke Elena and Klus, Stefan and Wu, Hao and Kutz, J Nathan and Brunton, Steven and Noé, Frank},
  journal={Machine Learning: Science and Technology},
  year={2021},
  publisher={IOP Publishing}
}

@article{mdtraj,
    title = {MDTraj: A Modern Open Library for the Analysis of
    Molecular Dynamics Trajectories},
    author = {McGibbon, Robert T. and Beauchamp, Kyle A. and Harrigan,
    Matthew P. and Klein, Christoph and Swails, Jason M. and
    Hern{\'a}ndez, Carlos X.  and Schwantes, Christian R. and Wang,
    Lee-Ping and Lane, Thomas J. and Pande, Vijay S.},
    journal = {Biophysical Journal},
    volume = {109},
    number = {8},
    pages = {1528 -- 1532},
    year = {2015},
    doi = {10.1016/j.bpj.2015.08.015}
}

@article{scikit-learn,
  title={Scikit-learn: Machine Learning in {P}ython},
  author={Pedregosa, F. and Varoquaux, G. and Gramfort, A. and Michel, V.
          and Thirion, B. and Grisel, O. and Blondel, M. and Prettenhofer, P.
          and Weiss, R. and Dubourg, V. and Vanderplas, J. and Passos, A. and
          Cournapeau, D. and Brucher, M. and Perrot, M. and Duchesnay, E.},
  journal={Journal of Machine Learning Research},
  volume={12},
  pages={2825--2830},
  year={2011}
}

@article{scipy,
  author  = {Virtanen, Pauli and Gommers, Ralf and Oliphant, Travis E. and
            Haberland, Matt and Reddy, Tyler and Cournapeau, David and
            Burovski, Evgeni and Peterson, Pearu and Weckesser, Warren and
            Bright, Jonathan and {van der Walt}, St{\'e}fan J. and
            Brett, Matthew and Wilson, Joshua and Millman, K. Jarrod and
            Mayorov, Nikolay and Nelson, Andrew R. J. and Jones, Eric and
            Kern, Robert and Larson, Eric and Carey, C J and
            Polat, {\.I}lhan and Feng, Yu and Moore, Eric W. and
            {VanderPlas}, Jake and Laxalde, Denis and Perktold, Josef and
            Cimrman, Robert and Henriksen, Ian and Quintero, E. A. and
            Harris, Charles R. and Archibald, Anne M. and
            Ribeiro, Ant{\^o}nio H. and Pedregosa, Fabian and
            {van Mulbregt}, Paul and {SciPy 1.0 Contributors}},
  title   = {{{SciPy} 1.0: Fundamental Algorithms for Scientific
            Computing in Python}},
  journal = {Nature Methods},
  year    = {2020},
  volume  = {17},
  pages   = {261--272},
  adsurl  = {https://rdcu.be/b08Wh},
  doi     = {10.1038/s41592-019-0686-2},
}

@article{deshawcovid,
  title={Molecular dynamics simulations related to Sars-Cov-2},
  author={Shaw, DE},
  journal={DE Shaw Research Technical Data},
  year={2020}
}

\clearpage
\renewcommand\thefigure{S\arabic{figure}}
\renewcommand\thetable{S\arabic{table}}
\renewcommand\theequation{S\arabic{equation}}
\renewcommand\thepage{S\arabic{page}}
\setcounter{figure}{0}
\setcounter{table}{0}
\setcounter{equation}{0}
\setcounter{page}{1}

\appendix

\begin{center}
\section*{
    Supplementary Materials for \\ 
    \emph{Representation choice shapes the interpretation of protein conformational dynamics}
}
Axel Giottonini$^{1,2}$, Thomas Lemmin$^{1,3}$
\begin{enumerate}
    \item \small Institute of Biochemistry and Molecular Medicine,
          University of Bern, Bühlstrasse 28, 3012 Bern, Switzerland
    \item \small Graduate School for Cellular and Biomedical Sciences (GCB),
          University of Bern, Mittelstrasse 43, 3012 Bern, Switzerland
    \item \small Department of Digital Medicine,
          University of Bern, Murtenstrasse 31, 3008 Bern
\end{enumerate}
\end{center}

\subsubsection*{This PDF file includes:}
Figures S1 to S23\\
Table S1 \\

\clearpage
\begin{figure}
    \includegraphics[width=\textwidth]{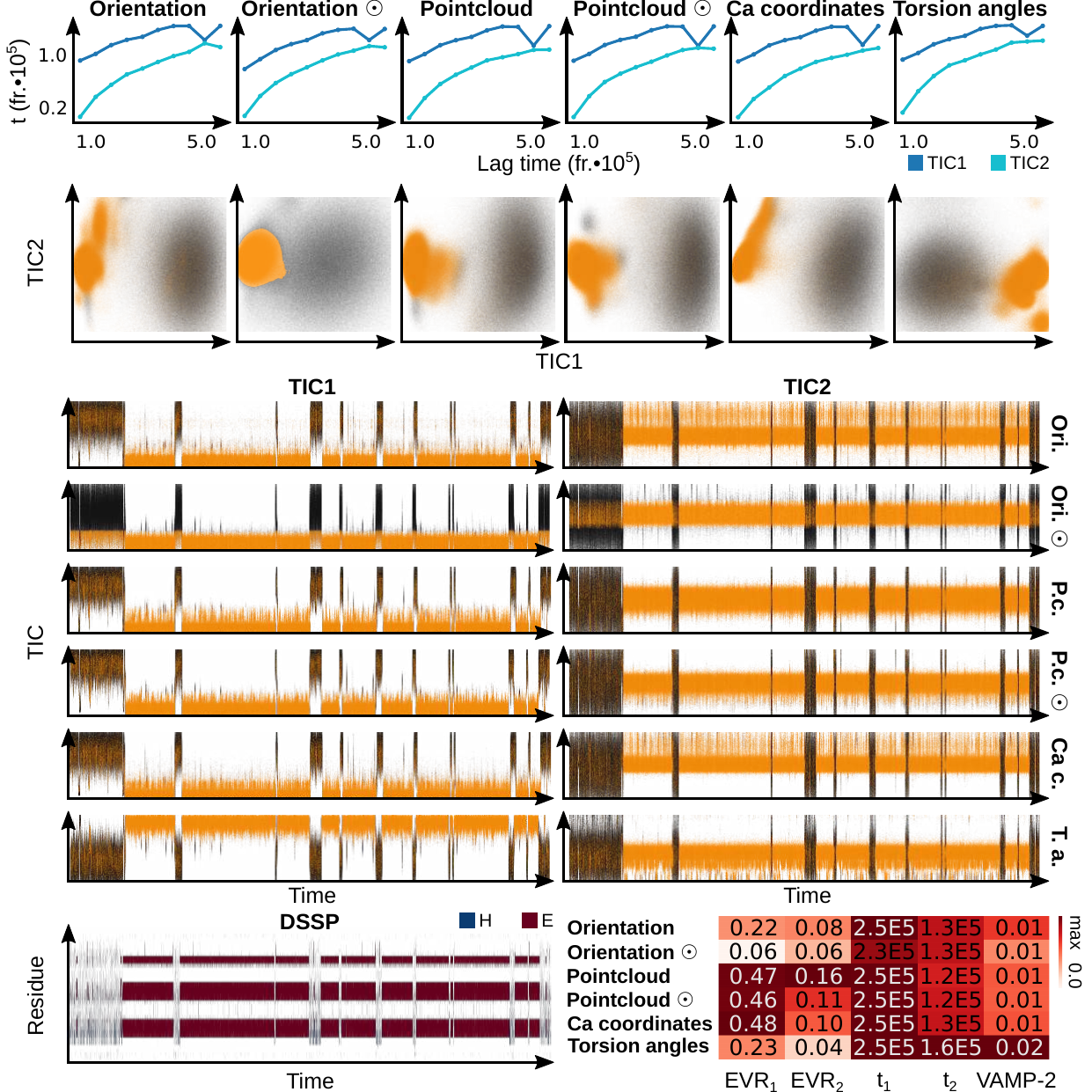}
    \caption{
    \textbf{WW Domain Analysis.}
    (\textbf{Row 1}) Implied timescales vs. lag time (log scale); dark/light blue: TIC1/TIC2.
    (\textbf{Row 2}) TIC1–TIC2 projections colored by Orientation~$\odot$ clusters (gray: outliers).
    (\textbf{Row 3}) TIC1 (\textit{left}) and TIC2 (\textit{right}) time series by cluster.
    (\textbf{Row 4}) DSSP heatmap (blue: $\alpha$-helix; red: $\beta$-sheet) (\textit{left}); column-normalized metrics heatmap (EVR1/2, t1/2, VAMP-2) (\textit{right}).
    }
    \label{fig:figure_s1}
\end{figure}

\begin{figure}
    \includegraphics[width=\textwidth]{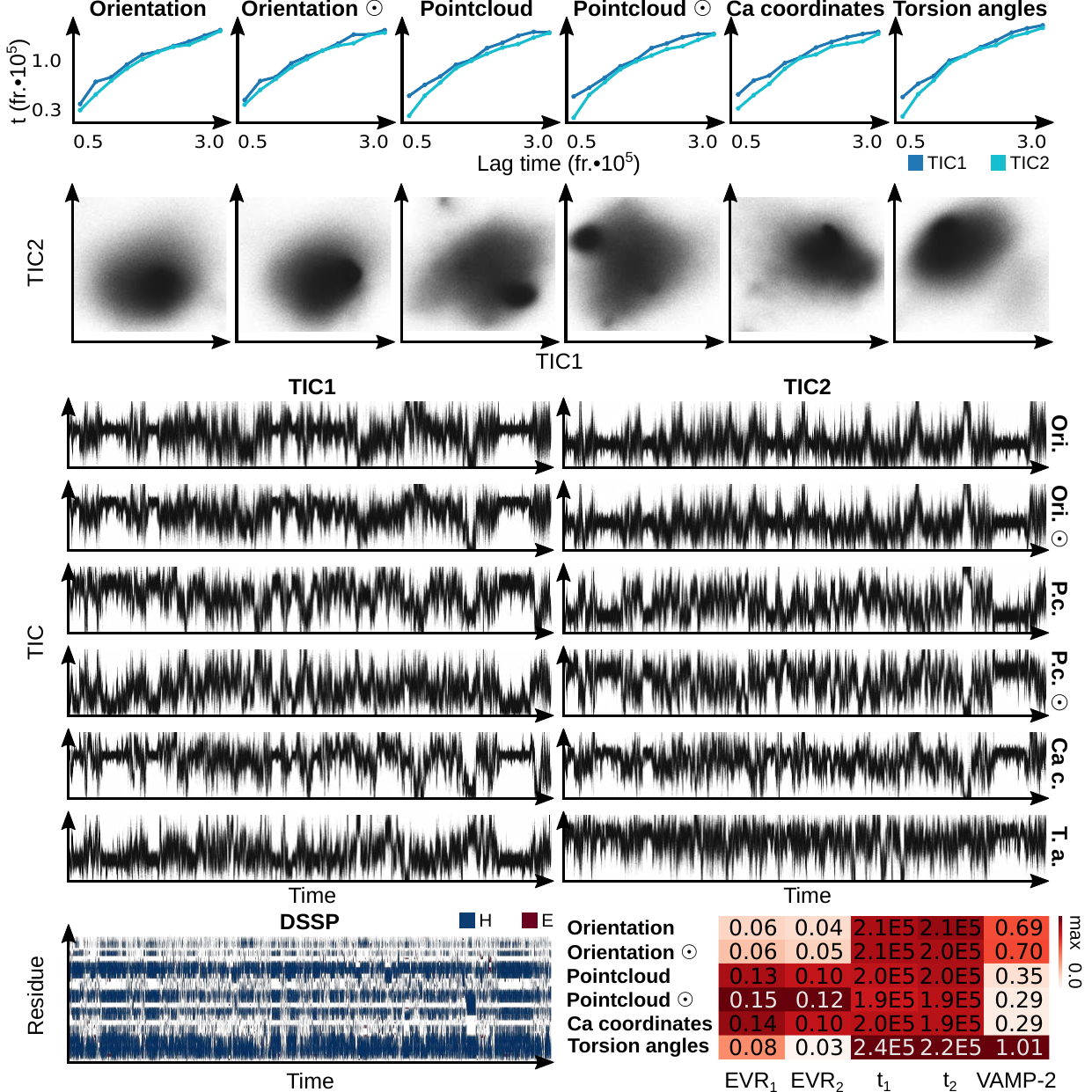}
    \caption{
    \textbf{$\lambda$-Repressor Analysis.}
    (\textbf{Row 1}) Implied timescales vs. lag time (log scale); dark/light blue: TIC1/TIC2.
    (\textbf{Row 2}) TIC1–TIC2 projections colored by Orientation~$\odot$ clusters (gray: outliers).
    (\textbf{Row 3}) TIC1 (\textit{left}) and TIC2 (\textit{right}) time series by cluster.
    (\textbf{Row 4}) DSSP heatmap (blue: $\alpha$-helix; red: $\beta$-sheet) (\textit{left}); column-normalized metrics heatmap (EVR1/2, t1/2, VAMP-2) (\textit{right}).
    }
    \label{fig:figure_s2}
\end{figure}

\begin{figure}
    \includegraphics[width=\textwidth]{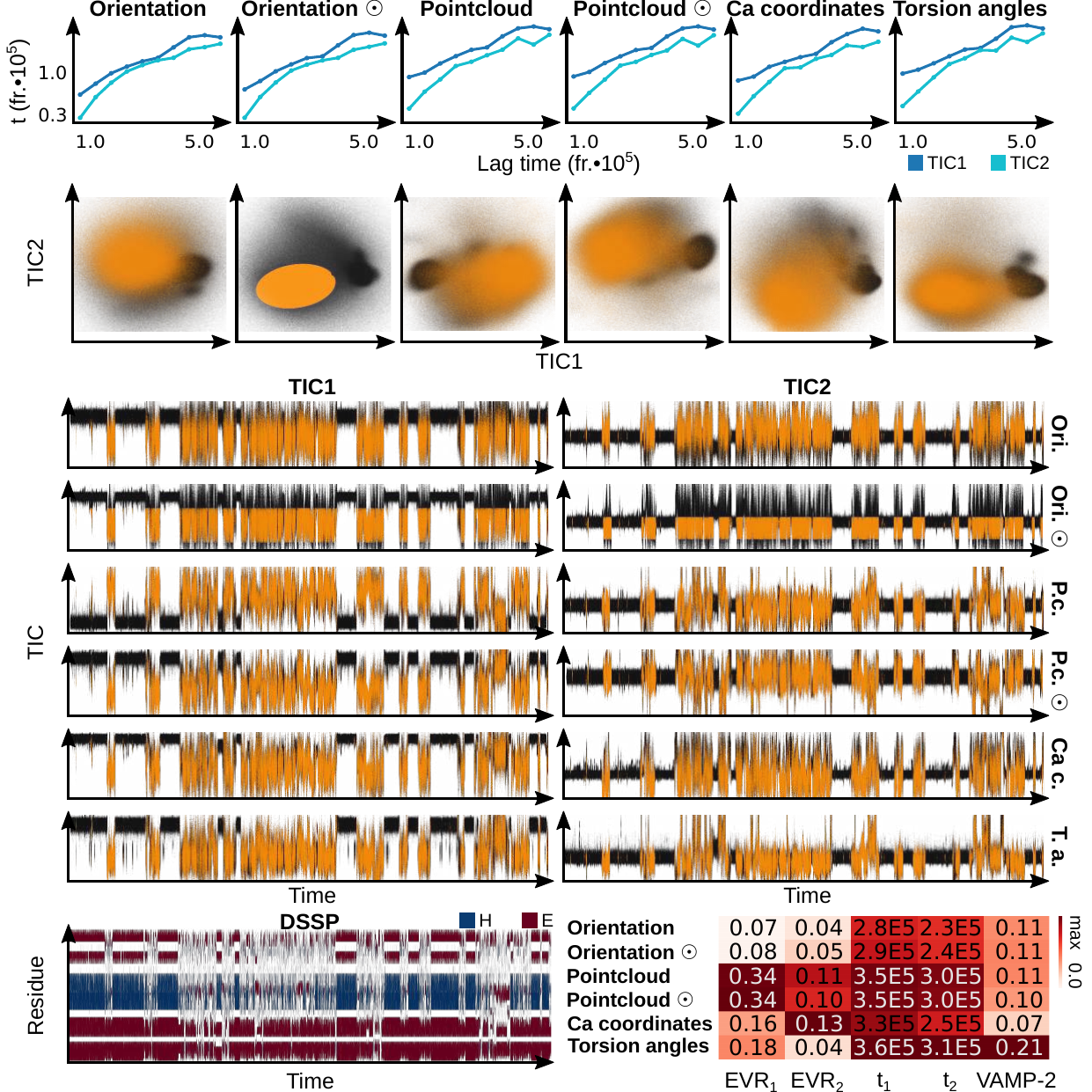}
    \caption{
    \textbf{Protein G Analysis.}
    (\textbf{Row 1}) Implied timescales vs. lag time (log scale); dark/light blue: TIC1/TIC2.
    (\textbf{Row 2}) TIC1–TIC2 projections colored by Orientation~$\odot$ clusters (gray: outliers).
    (\textbf{Row 3}) TIC1 (\textit{left}) and TIC2 (\textit{right}) time series by cluster.
    (\textbf{Row 4}) DSSP heatmap (blue: $\alpha$-helix; red: $\beta$-sheet) (\textit{left}); column-normalized metrics heatmap (EVR1/2, t1/2, VAMP-2) (\textit{right}).
    }
    \label{fig:figure_s3}
\end{figure}

\begin{figure}
    \includegraphics[width=\textwidth]{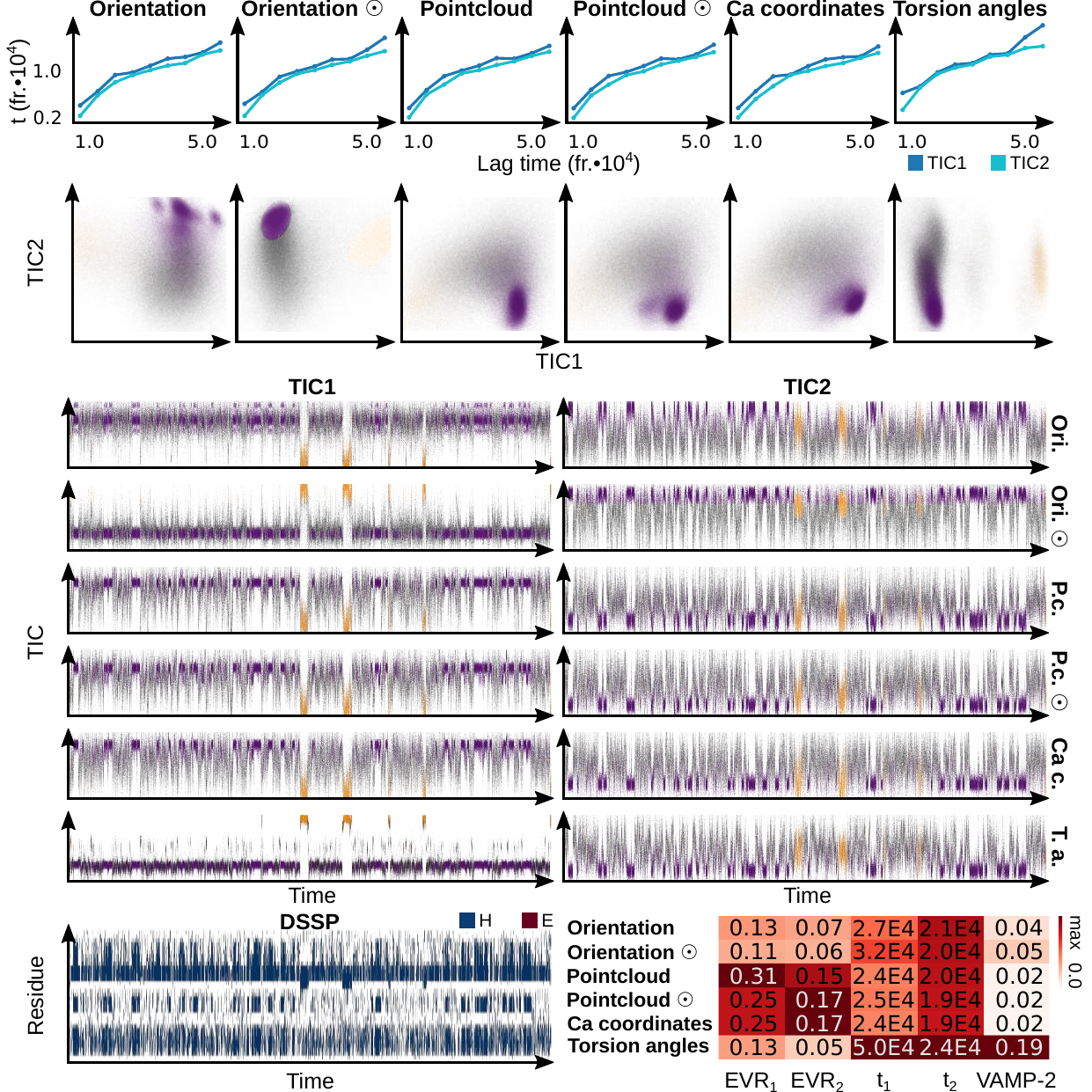}
    \caption{
    \textbf{Villin Analysis.}
    (\textbf{Row 1}) Implied timescales vs. lag time (log scale); dark/light blue: TIC1/TIC2.
    (\textbf{Row 2}) TIC1–TIC2 projections colored by Orientation~$\odot$ clusters (gray: outliers).
    (\textbf{Row 3}) TIC1 (\textit{left}) and TIC2 (\textit{right}) time series by cluster.
    (\textbf{Row 4}) DSSP heatmap (blue: $\alpha$-helix; red: $\beta$-sheet) (\textit{left}); column-normalized metrics heatmap (EVR1/2, t1/2, VAMP-2) (\textit{right}).
    }
    \label{fig:figure_s4}
\end{figure}

\begin{figure}
    \includegraphics[width=\textwidth]{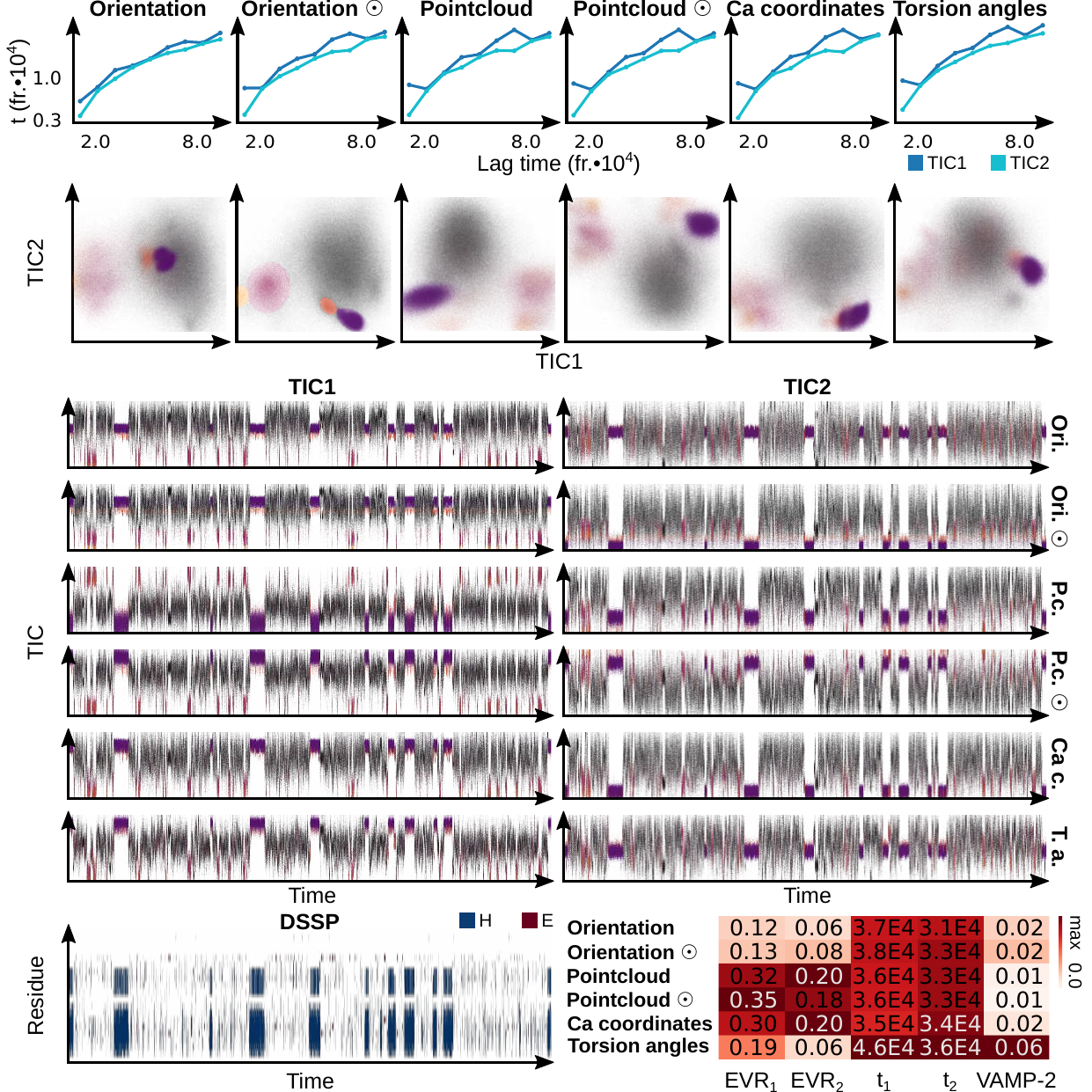}
    \caption{
    \textbf{Trp-Cage Analysis.}
    (\textbf{Row 1}) Implied timescales vs. lag time (log scale); dark/light blue: TIC1/TIC2.
    (\textbf{Row 2}) TIC1–TIC2 projections colored by Orientation~$\odot$ clusters (gray: outliers).
    (\textbf{Row 3}) TIC1 (\textit{left}) and TIC2 (\textit{right}) time series by cluster.
    (\textbf{Row 4}) DSSP heatmap (blue: $\alpha$-helix; red: $\beta$-sheet) (\textit{left}); column-normalized metrics heatmap (EVR1/2, t1/2, VAMP-2) (\textit{right}).
    }
    \label{fig:figure_s5}
\end{figure}

\begin{figure}
    \includegraphics[width=\textwidth]{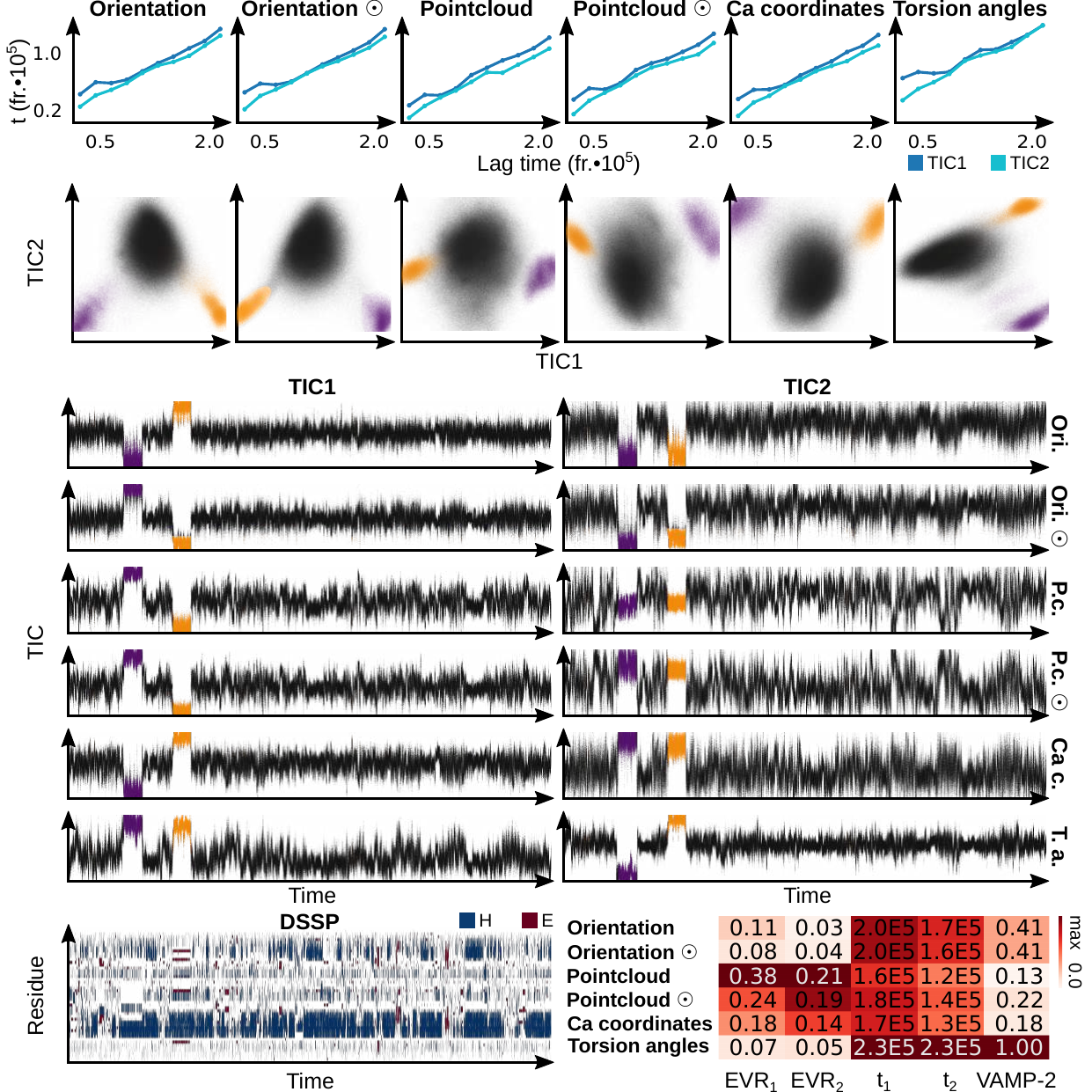}
    \caption{
    \textbf{BBL Analysis.}
    (\textbf{Row 1}) Implied timescales vs. lag time (log scale); dark/light blue: TIC1/TIC2.
    (\textbf{Row 2}) TIC1–TIC2 projections colored by Orientation~$\odot$ clusters (gray: outliers).
    (\textbf{Row 3}) TIC1 (\textit{left}) and TIC2 (\textit{right}) time series by cluster.
    (\textbf{Row 4}) DSSP heatmap (blue: $\alpha$-helix; red: $\beta$-sheet) (\textit{left}); column-normalized metrics heatmap (EVR1/2, t1/2, VAMP-2) (\textit{right}).
    }
    \label{fig:figure_s6}
\end{figure}

\begin{figure}
    \includegraphics[width=\textwidth]{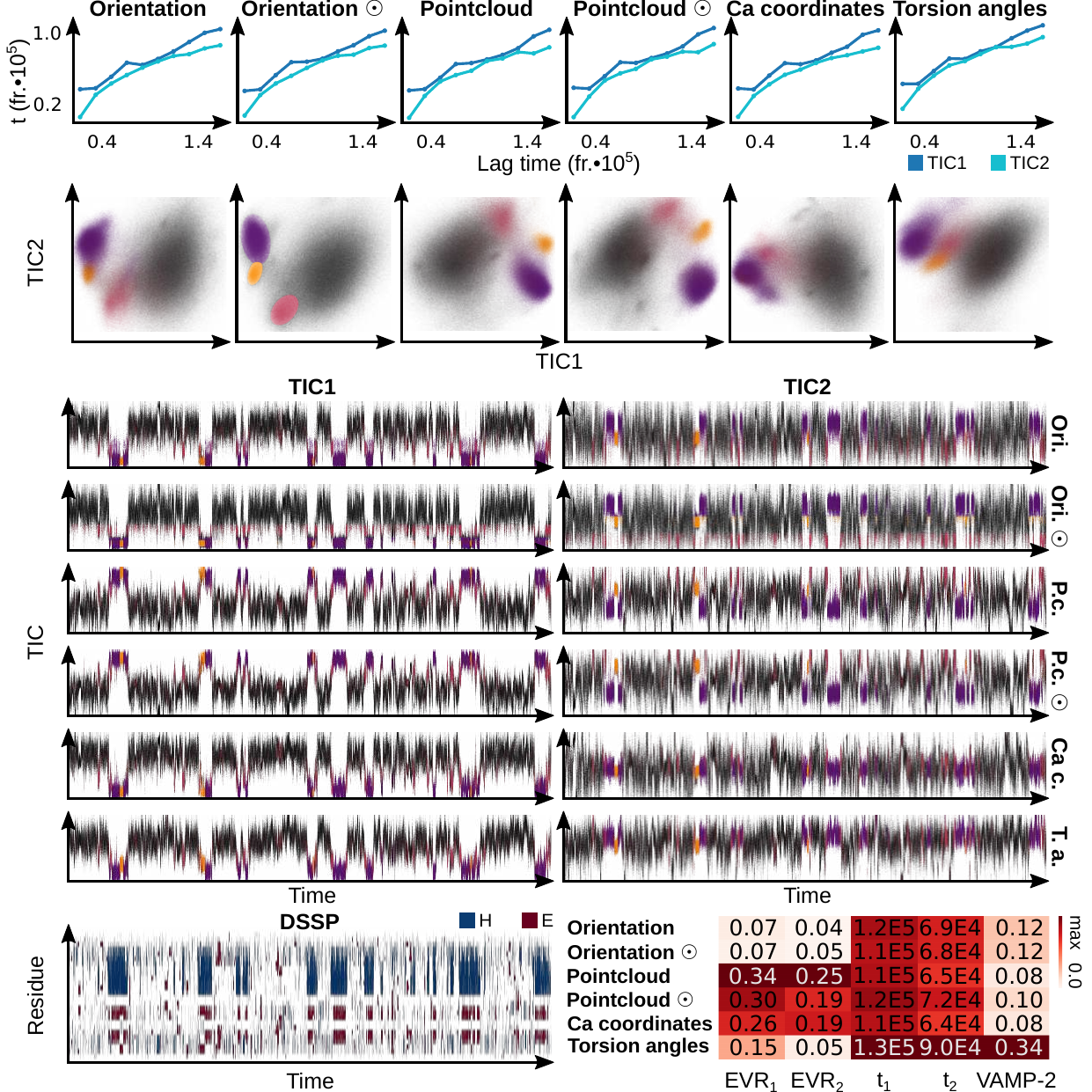}
    \caption{
    \textbf{BBA Analysis.}
    (\textbf{Row 1}) Implied timescales vs. lag time (log scale); dark/light blue: TIC1/TIC2.
    (\textbf{Row 2}) TIC1–TIC2 projections colored by Orientation~$\odot$ clusters (gray: outliers).
    (\textbf{Row 3}) TIC1 (\textit{left}) and TIC2 (\textit{right}) time series by cluster.
    (\textbf{Row 4}) DSSP heatmap (blue: $\alpha$-helix; red: $\beta$-sheet) (\textit{left}); column-normalized metrics heatmap (EVR1/2, t1/2, VAMP-2) (\textit{right}).
    }
    \label{fig:figure_s7}
\end{figure}

\begin{figure}
    \includegraphics[width=\textwidth]{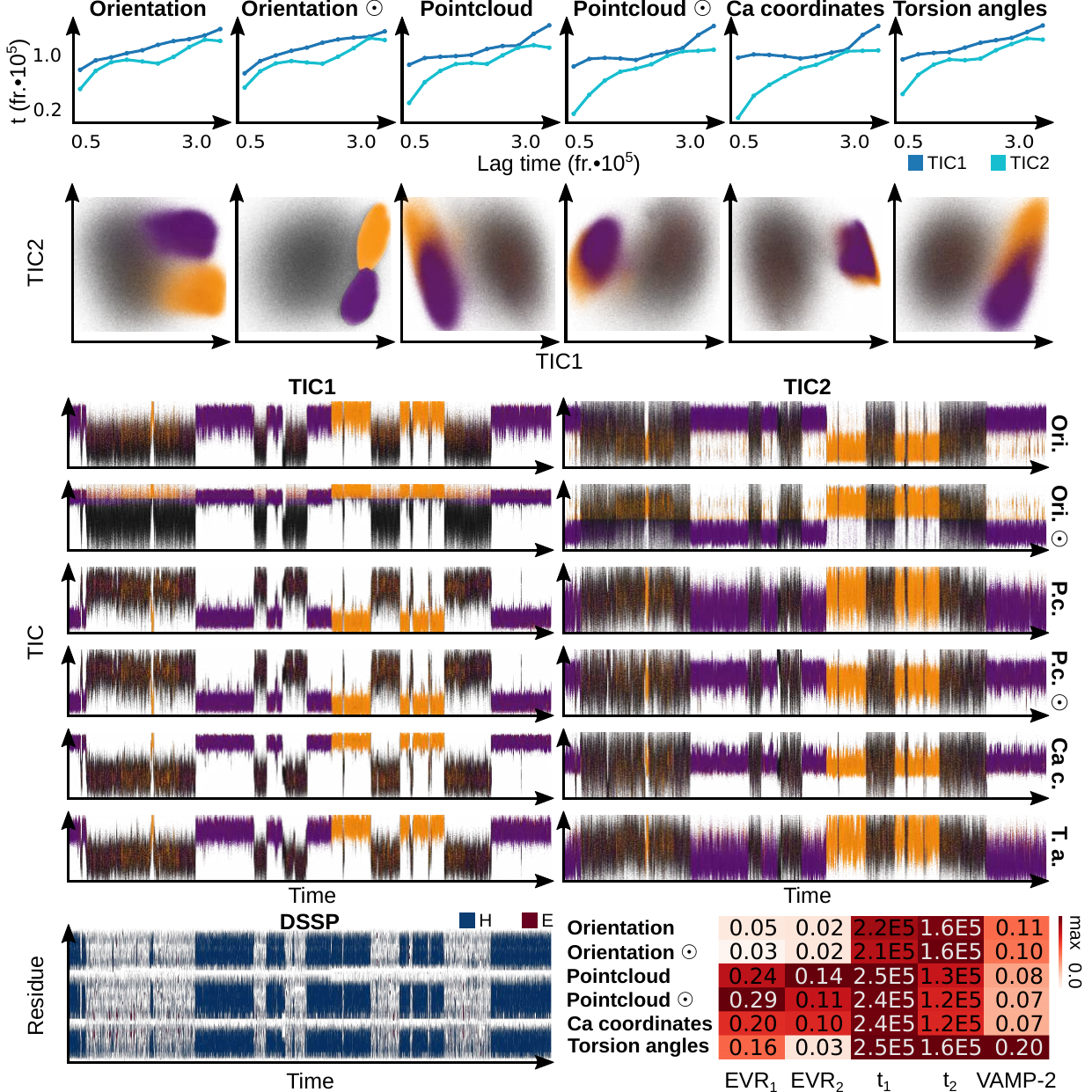}
    \caption{
    \textbf{$\alpha$3D Analysis.}
    (\textbf{Row 1}) Implied timescales vs. lag time (log scale); dark/light blue: TIC1/TIC2.
    (\textbf{Row 2}) TIC1–TIC2 projections colored by Orientation~$\odot$ clusters (gray: outliers).
    (\textbf{Row 3}) TIC1 (\textit{left}) and TIC2 (\textit{right}) time series by cluster.
    (\textbf{Row 4}) DSSP heatmap (blue: $\alpha$-helix; red: $\beta$-sheet) (\textit{left}); column-normalized metrics heatmap (EVR1/2, t1/2, VAMP-2) (\textit{right}).
    }
    \label{fig:figure_s8}
\end{figure}

\begin{figure}
    \includegraphics[width=\textwidth]{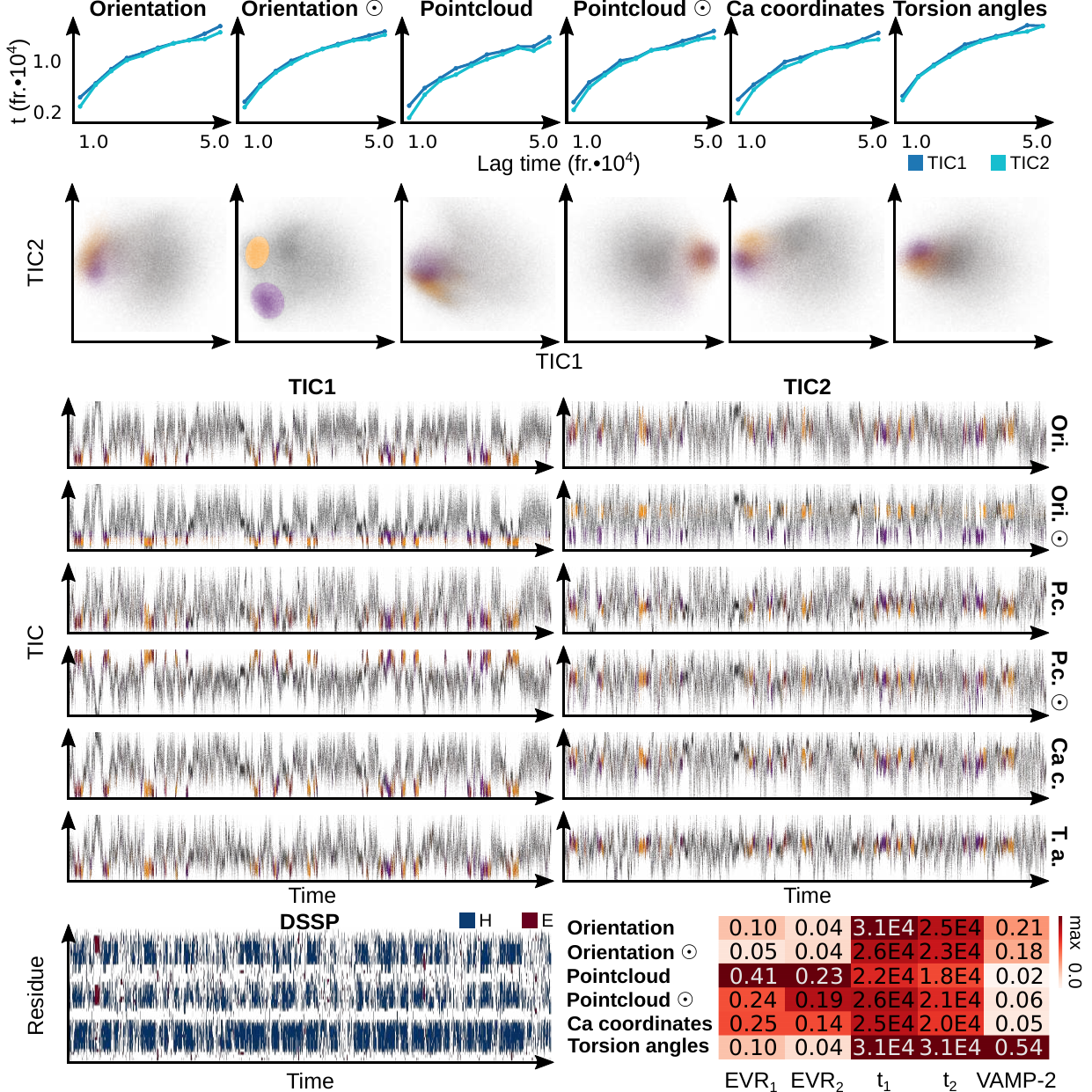}
    \caption{
    \textbf{Protein B Analysis.}
    (\textbf{Row 1}) Implied timescales vs. lag time (log scale); dark/light blue: TIC1/TIC2.
    (\textbf{Row 2}) TIC1–TIC2 projections colored by Orientation~$\odot$ clusters (gray: outliers).
    (\textbf{Row 3}) TIC1 (\textit{left}) and TIC2 (\textit{right}) time series by cluster.
    (\textbf{Row 4}) DSSP heatmap (blue: $\alpha$-helix; red: $\beta$-sheet) (\textit{left}); column-normalized metrics heatmap (EVR1/2, t1/2, VAMP-2) (\textit{right}).
    }
    \label{fig:figure_s9}
\end{figure}

\begin{figure}
    \includegraphics[width=\textwidth]{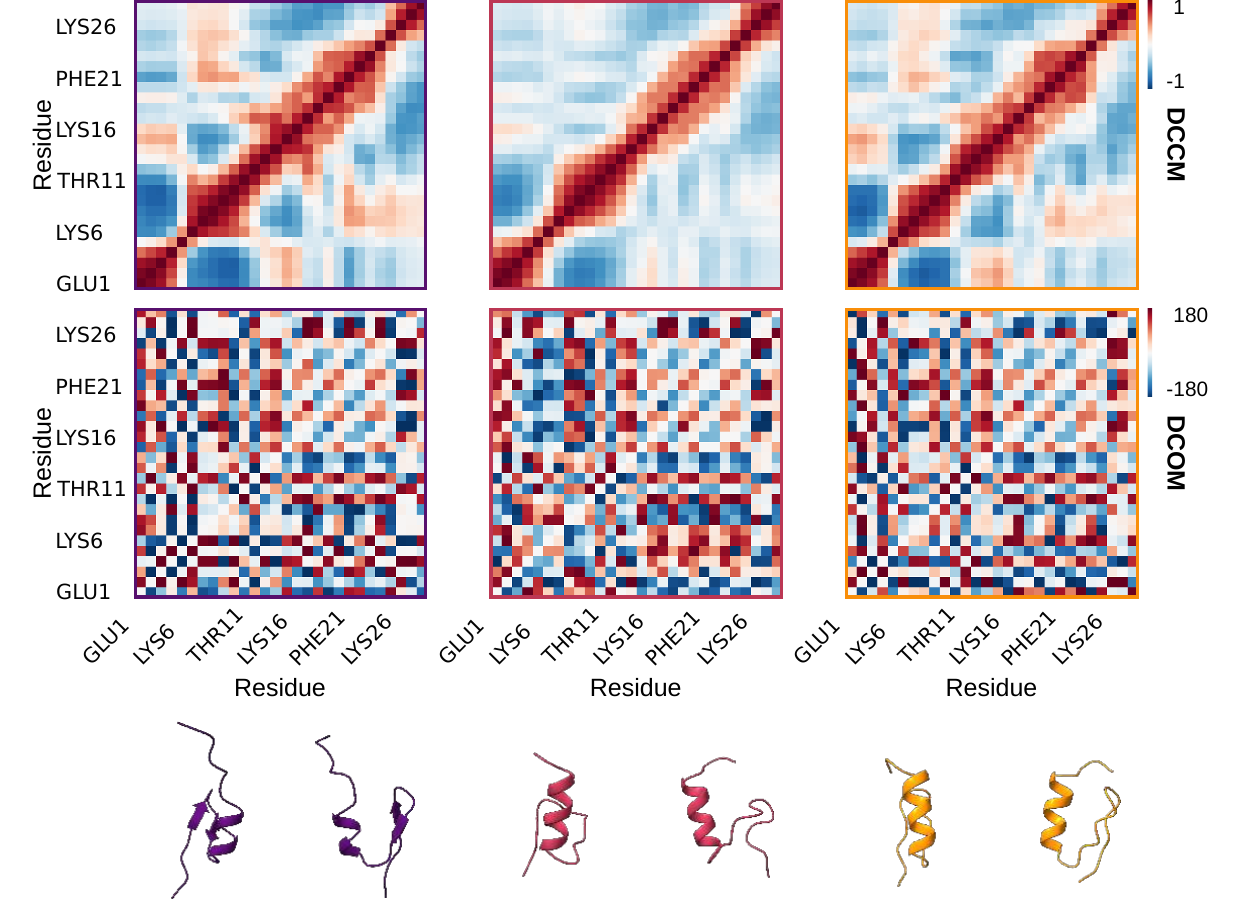}
    \caption{
    \textbf{BBA Clusters.}
    (\textbf{Top}) DCCM heatmaps (red: correlation; blue: anti-correlation).
    (\textbf{Middle}) DCOM heatmaps (red/blue: $\pm 180 \deg$ between residue normal vectors).
    (\textbf{Bottom}) Cluster medoid structures. Colors match Supplementary Figure~\ref{fig:figure_s7}.
    }
    \label{fig:figure_s10}
\end{figure}

\begin{figure}
    \includegraphics[width=\textwidth]{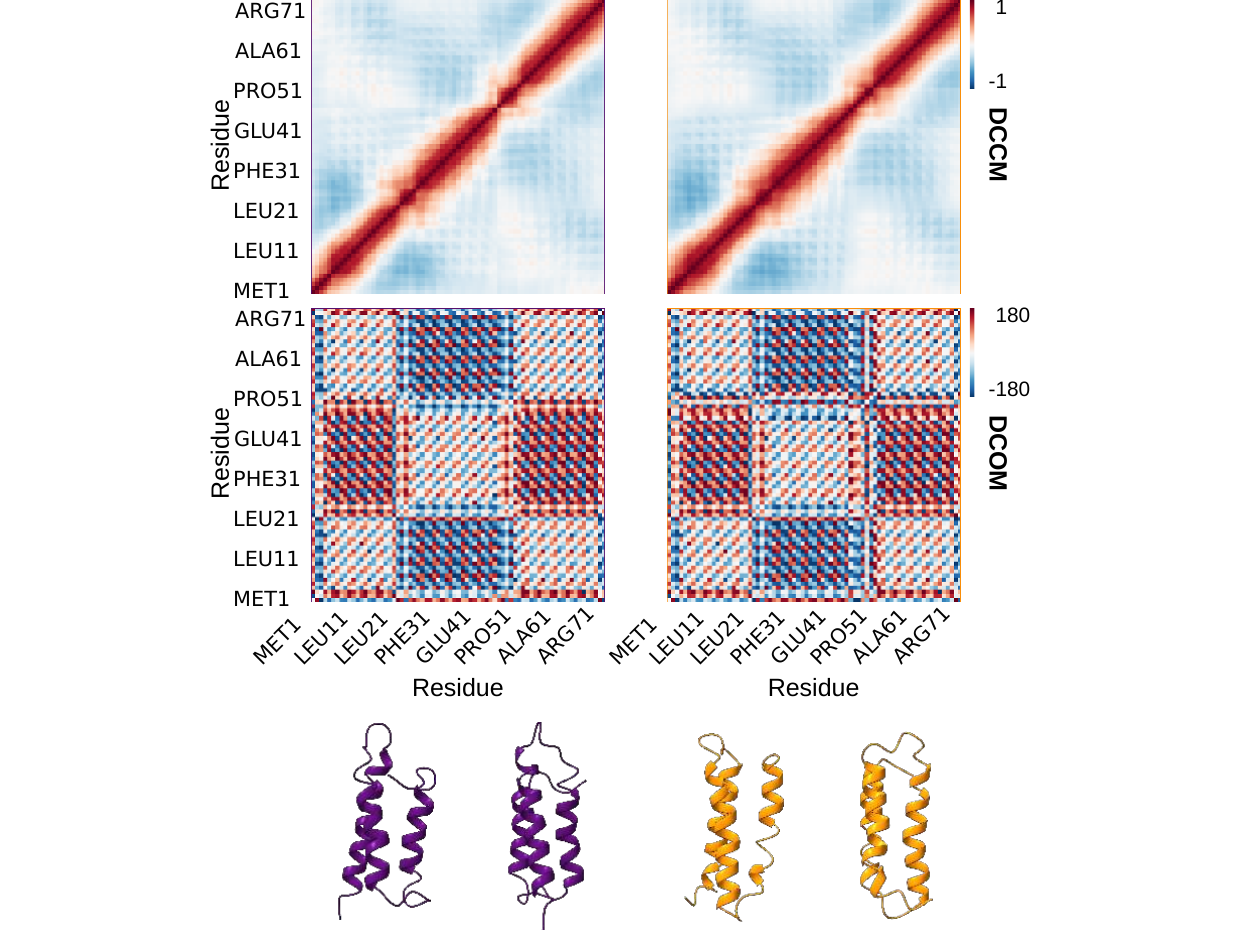}
    \caption{
    \textbf{$\alpha$3D Clusters.}
    (\textbf{Top}) DCCM heatmaps (red: correlation; blue: anti-correlation).
    (\textbf{Middle}) DCOM heatmaps (red/blue: $\pm 180 \deg$ between residue normal vectors).
    (\textbf{Bottom}) Cluster medoid structures. Colors match Supplementary Figure~\ref{fig:figure_s8}.
    }
    \label{fig:figure_s11}
\end{figure}

\begin{figure}
    \includegraphics[width=\textwidth]{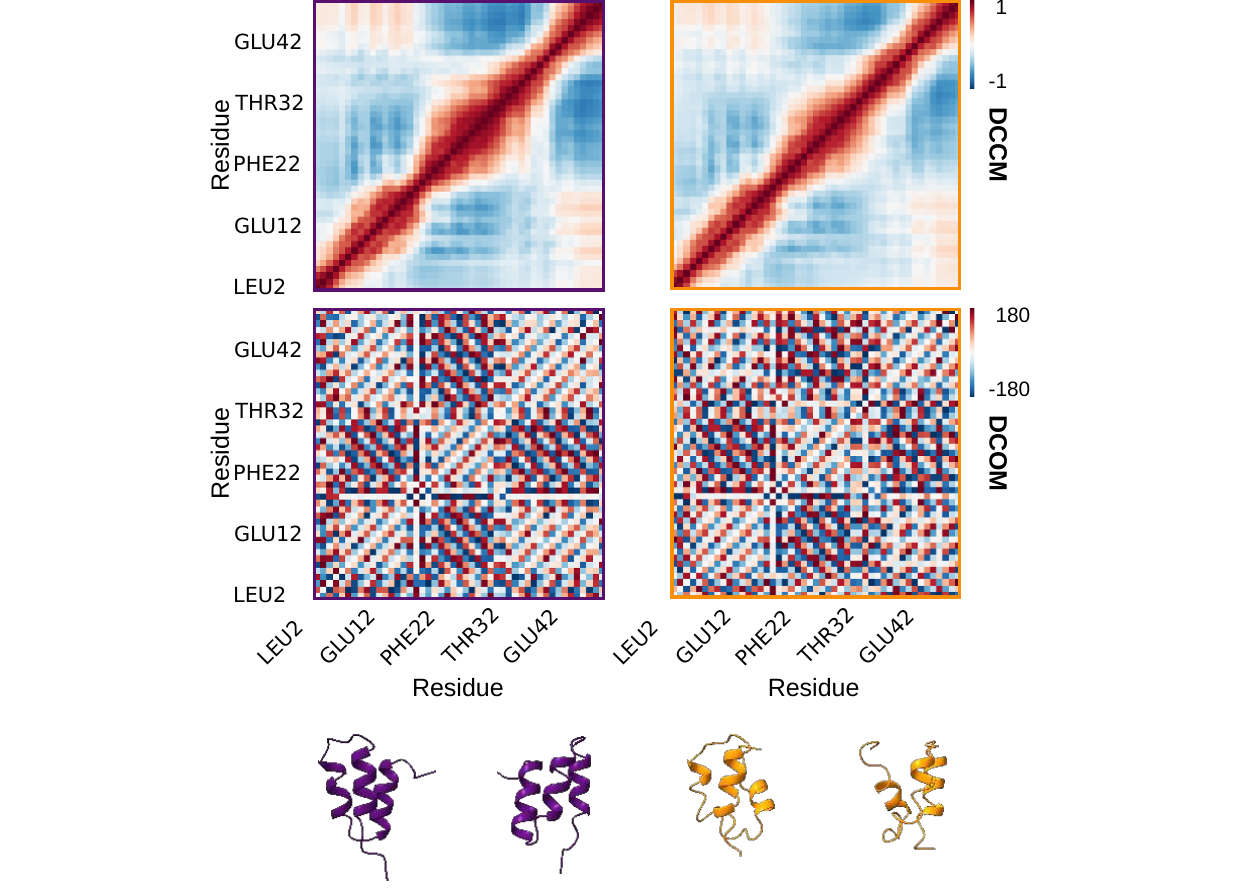}
    \caption{
    \textbf{Protein B Clusters.}
    (\textbf{Top}) DCCM heatmaps (red: correlation; blue: anti-correlation).
    (\textbf{Middle}) DCOM heatmaps (red/blue: $\pm 180 \deg$ between residue normal vectors).
    (\textbf{Bottom}) Cluster medoid structures. Colors match Supplementary Figure~\ref{fig:figure_s9}.
    }
    \label{fig:figure_s12}
\end{figure}

\begin{figure}
    \includegraphics[width=\textwidth]{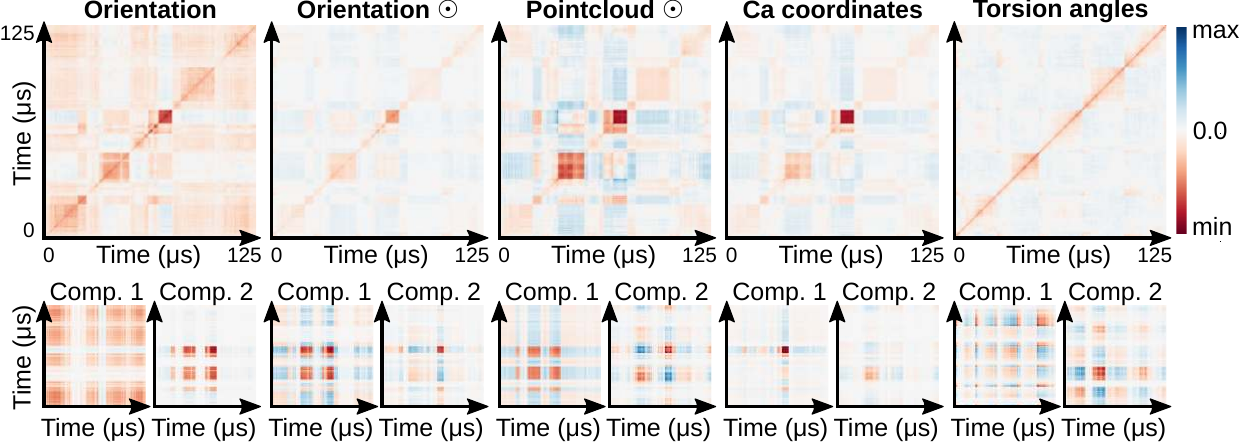}
    \caption{
    \textbf{nsp13 Gram Matrices.}
    (\textbf{Top}) Gram matrices for each representation across merged trajectories.
    (\textbf{Bottom}) First two rank-1 components from eigendecomposition.
    }
    \label{fig:figure_s13}
\end{figure}

\begin{figure}
    \includegraphics[width=\textwidth]{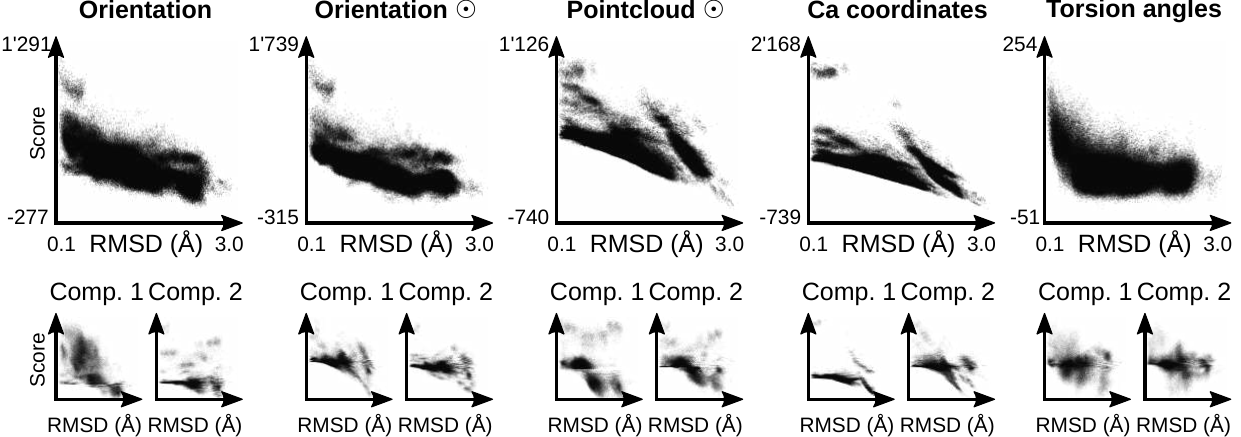}
    \caption{
    \textbf{nsp13 Gram Matrices - RMSD Correlation.}
    Scatterplots of Gram matrix scores vs. RMSD.
    (\textbf{Top}) Full Gram matrices.
    (\textbf{Bottom}) Rank-1 components.
    }
    \label{fig:figure_s14}
\end{figure}

\begin{figure}
    \includegraphics[width=\textwidth]{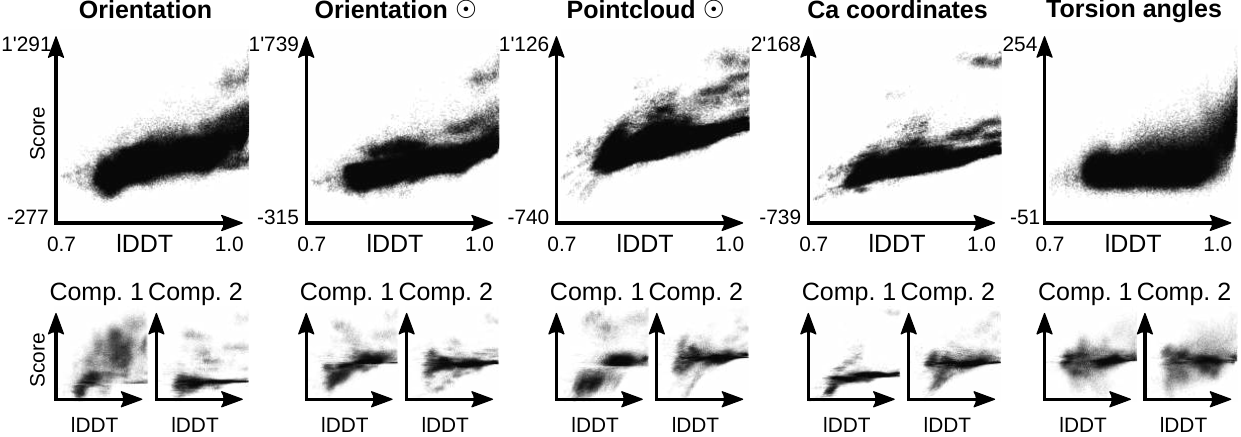}
    \caption{
    \textbf{nsp13 Gram Matrices - lDDT Correlation.}
    Scatterplots of Gram matrix scores vs. lDDT.
    (\textbf{Top}) Full Gram matrices.
    (\textbf{Bottom}) Rank-1 components.
    }
    \label{fig:figure_s15}
\end{figure}

\begin{figure}
    \includegraphics[width=\textwidth]{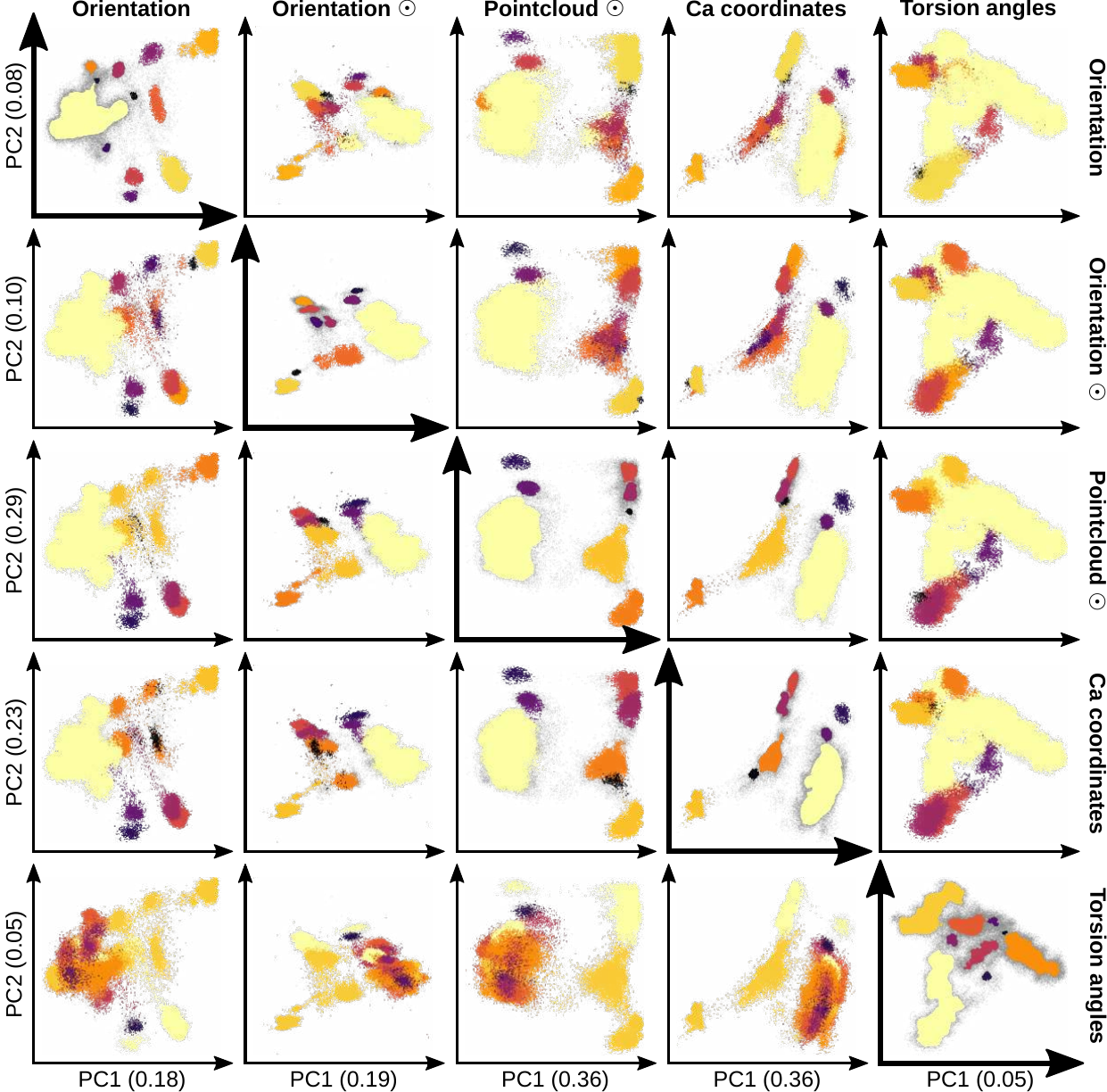}
    \caption{
    \textbf{nsp13 Clusters.}
    PC1–PC2 projections for each representation, colored by cluster assignment. Rows correspond to different clustering assignments.
    }
    \label{fig:figure_s16}
\end{figure}

\begin{figure}
    \includegraphics[width=\textwidth]{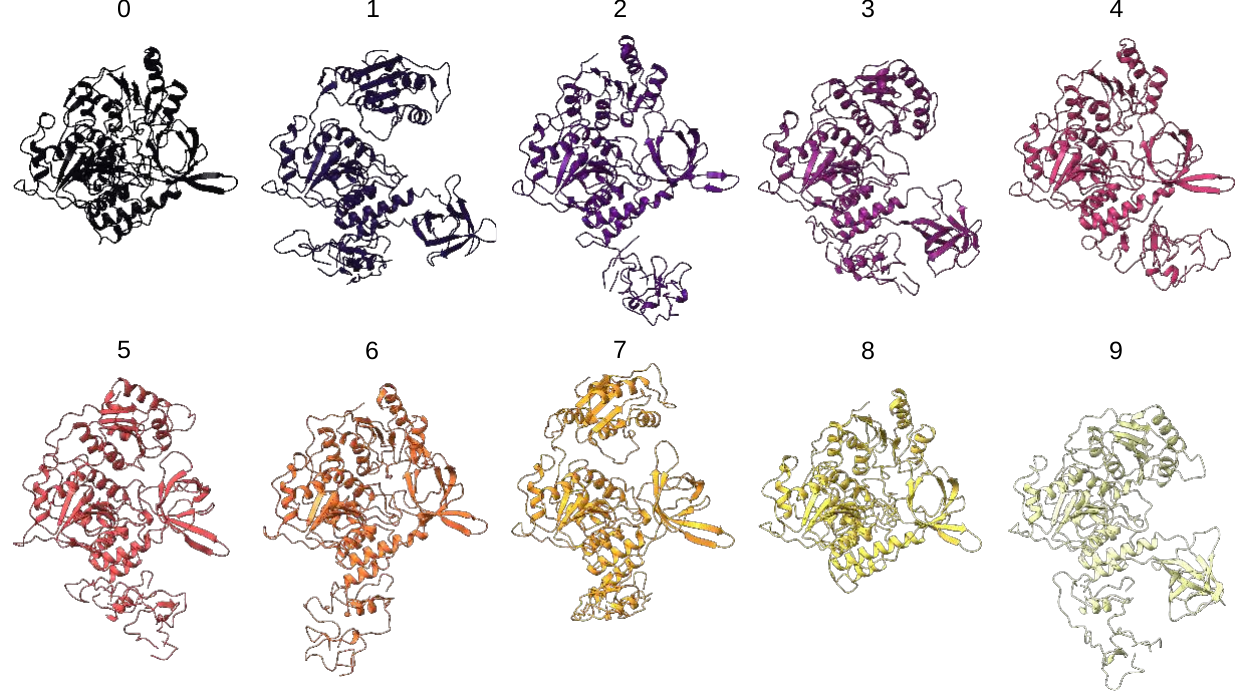}
    \caption{
    \textbf{nsp13 Clusters Structures.}
    Orientation~$\odot$ cluster centroids aligned on stalk and RecA1 domains. Colors match Supplementary Figure~\ref{fig:figure_s16}.
    }
    \label{fig:figure_s17}
\end{figure}

\begin{figure}
    \includegraphics[width=\textwidth]{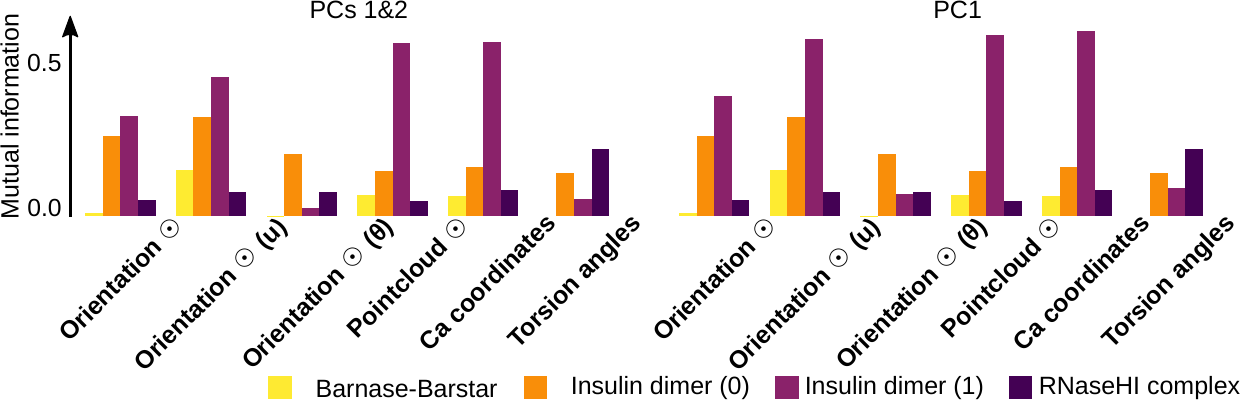}
    \caption{
    \textbf{Protein-Protein Association Mutual Information.}
    MI between association state and PC1–PC2 (\textit{left}) or PC1 alone (\textit{right}) for each representation. Colors indicate systems.
    }
    \label{fig:figure_s18}
\end{figure}

\begin{figure}
    \includegraphics[width=\textwidth]{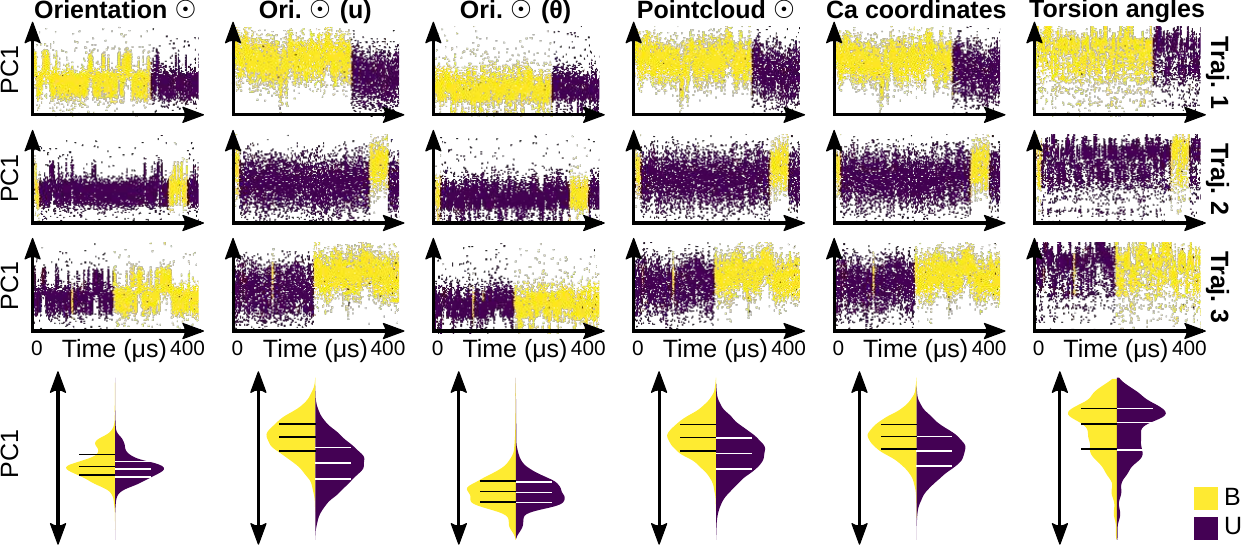}
    \caption{
    \textbf{Barnase-Barstar PC1 Distribution.}
    (\textbf{Top}) PC1 time series for each representation.
    (\textbf{Bottom}) Violin plots of PC1 by association state (B: bound; U: unbound); lines indicate quartiles.
    }
    \label{fig:figure_s19}
\end{figure}

\begin{figure}
    \includegraphics[width=\textwidth]{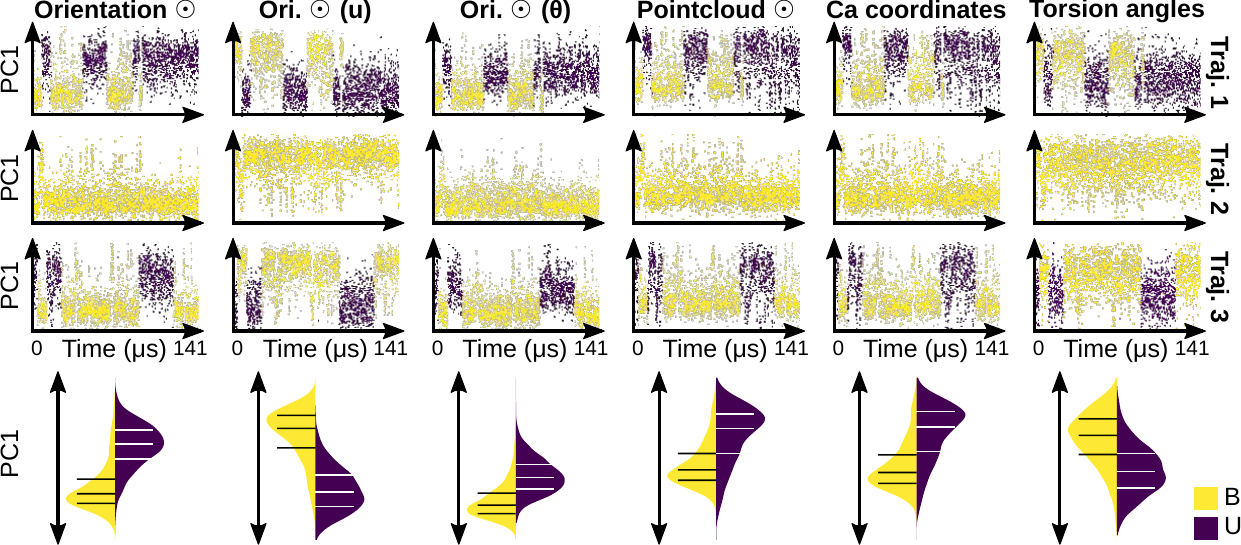}
    \caption{
    \textbf{Insulin Dimer (0) PC1 Distribution.}
    (\textbf{Top}) PC1 time series for each representation.
    (\textbf{Bottom}) Violin plots of PC1 by association state (B: bound; U: unbound); lines indicate quartiles.
    }
    \label{fig:figure_s20}
\end{figure}

\begin{figure}
    \includegraphics[width=\textwidth]{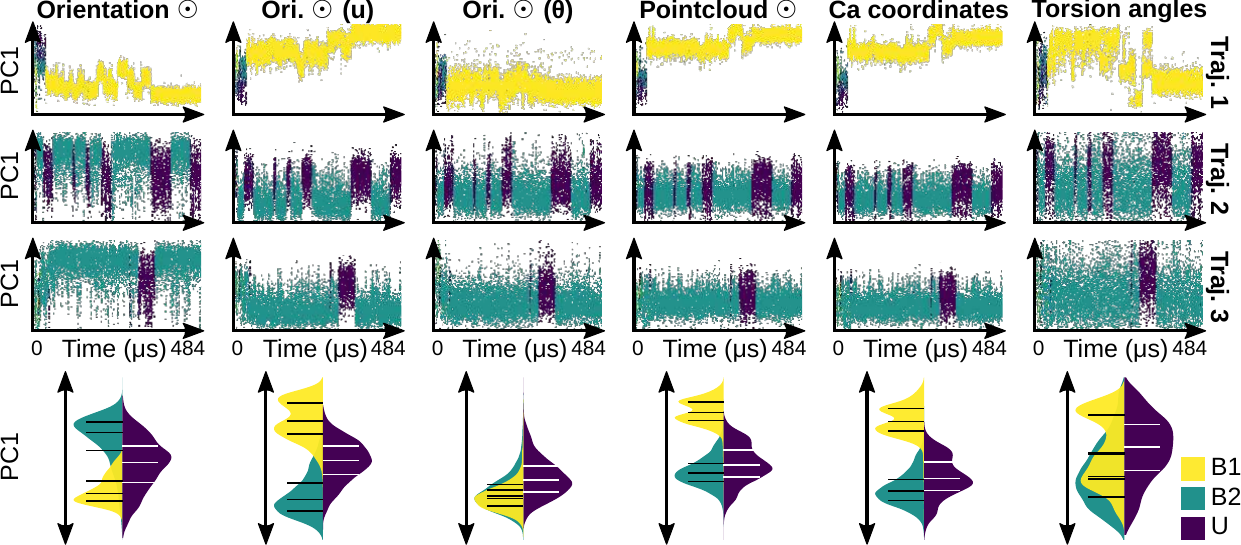}
    \caption{
    \textbf{Insulin Dimer (1) PC1 Distribution.}
    (\textbf{Top}) PC1 time series for each representation.
    (\textbf{Bottom}) Violin plots of PC1 by association state (B: bound; B': non-canonical bound; U: unbound); lines indicate quartiles.
    }
    \label{fig:figure_s21}
\end{figure}

\begin{figure}
    \includegraphics[width=\textwidth]{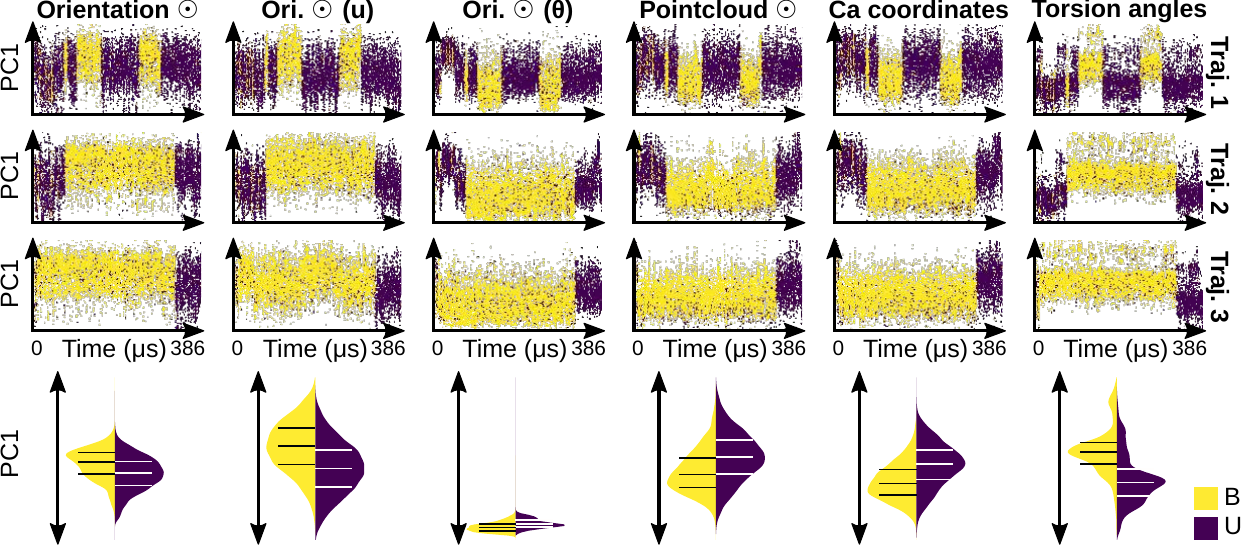}
    \caption{
    \textbf{RNaseHI-SSB Complex PC1 Distribution.}
    (\textbf{Top}) PC1 time series for each representation.
    (\textbf{Bottom}) Violin plots of PC1 by association state (B: bound; U: unbound); lines indicate quartiles.
    }
    \label{fig:figure_s22}
\end{figure}

\begin{figure}
    \includegraphics[width=\textwidth]{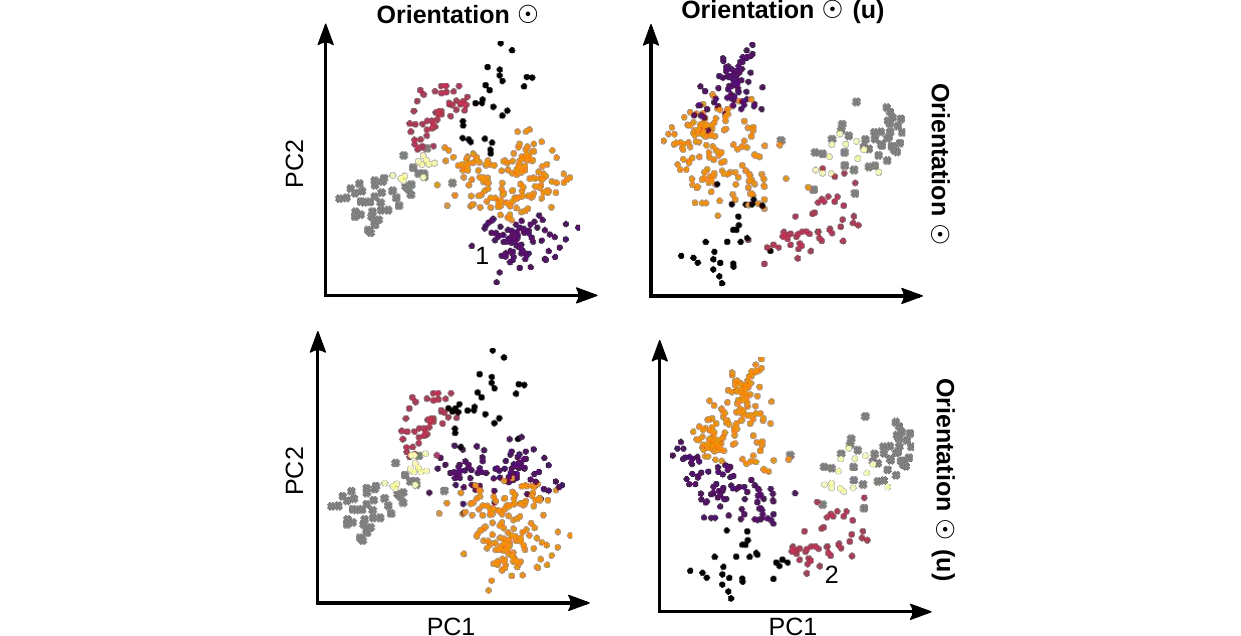}
    \caption{
    \textbf{RBD-ACE2 Clusters.}
    PC1–PC2 projections for each representation, colored by cluster assignment. Rows correspond to different clustering solutions.
    }
    \label{fig:figure_s23}
\end{figure}

\clearpage
\begin{table}
	\centering
	\caption{
    \textbf{Simulation Details.}
    Number of frames --total and analyzed--, subsampling stride and time in $\mu$s of the systems analyzed.
    }
	\label{tab:table_s1}

	\begin{tabular}{lrrrr}
		\\
		\hline
		System & N. Frames (total) & N. Frames (analyzed) & Stride & Time~($\mu s$) \\
		\hline
        BBA & 1'625'195 & 1'625'195 & 1 & 325.0 \\
        Villin & 627'907 & 627'907 & 1 & 125.0 \\
        Trp-cage & 1'044'000 & 1'044'000 & 1 & 208.0 \\
        BBL & 2'150'146 & 2'150'146 & 1 & 429.0 \\
        $\alpha$3D & 3'535'447 & 3'535'447 & 1 & 707.0 \\
        WW domain & 5'686'712 & 5'686'712 & 1 & 1'137.0 \\
        $\lambda$-repressor & 3'235'363 & 3'235'363 & 1 & 643.0 \\
        Protein G & 5'781'376 & 5'781'376 & 1 & 1154.0 \\
        Protein B & 520'250 & 520'250 & 1 & 104.0 \\
        SARS-CoV-2 nsp13 & 104'170 & 104'170 & 1 & 5 $\times$ 25.0 \\
        Barnase–Barstar & 5'849'318 & 22'812 & 256 & 3 $\times$ 400.0 \\
        Insulin dimer (0) & 2'065'624 & 8'055 & 256 & 3 $\times$ 140.7 \\
        Insulin dimer (1) & 7'079'495 & 27'610 & 256 & 3 $\times$ 484.2 \\
        RNaseHI-SSB complex & 6'299'957 & 24'569 & 256 & 3 $\times$ 385.6 \\
        RBD-ACE2 & 618 & 618 & 1 & - \\
	\end{tabular}
\end{table}

\end{document}